\documentclass{article}

\PassOptionsToPackage{numbers,sort&compress}{natbib}

\usepackage[eandd, preprint]{neurips_2026}

\usepackage{siunitx}
\usepackage[utf8]{inputenc} 
\usepackage[T1]{fontenc}    
\usepackage[pagebackref,breaklinks,colorlinks]{hyperref} 
\usepackage{url}            
\usepackage{booktabs}       
\usepackage{amsfonts}       
\usepackage{nicefrac}       
\usepackage{microtype}      
\usepackage{xcolor}         
\usepackage{colortbl}       
\usepackage{tabularx}       
\definecolor{tabblue}{HTML}{1a6fb3}    
\definecolor{tabred}{HTML}{c2185b}     
\definecolor{tabamber}{HTML}{b36a1a}   
\definecolor{tabpurple}{HTML}{5e35b1}  
\usepackage{pifont}
\usepackage{paralist}
\usepackage{xspace}
\usepackage{graphicx}
\usepackage{subcaption}
\usepackage{float}
\usepackage{placeins}           
\usepackage{etoolbox}
\newif\ifappendixphase
\appendixphasefalse
\AtBeginEnvironment{subsection}{\ifappendixphase\FloatBarrier\fi}
\AtBeginEnvironment{section}{\ifappendixphase\FloatBarrier\fi}
\usepackage{multirow}
\usepackage[export]{adjustbox}
\usepackage{caption}
\usepackage{array}
\usepackage{courier}
\usepackage{makecell}
\usepackage{tikz}
\usetikzlibrary{positioning}

\usepackage[most]{tcolorbox}
\tcbuselibrary{listings,skins,breakable}
\usepackage{listings}
\usepackage{helvet}
\usepackage{cleveref}

\lstdefinelanguage{json}{
  basicstyle=\ttfamily\footnotesize,
  numbers=none, showstringspaces=false, breaklines=true,
  morestring=[s]{"}{"}, morecomment=[l]{:\ },
}

\definecolor{ForestGreen}{rgb}{0.13, 0.55, 0.13}

\newcommand{\model}[1]{{{\small\fontfamily{phv}\selectfont{#1}}\xspace}}
\definecolor{gpt_green}{RGB}{22,163,127}
\definecolor{gemini_blue}{RGB}{81,134,209}
\definecolor{claude_orange}{RGB}{216,119,87}
\definecolor{qwen_violet}{RGB}{191,123,234}
\definecolor{pixtral_orange}{RGB}{255,138,0}
\definecolor{llama_blue}{RGB}{0,102,204}
\definecolor{internvl_blue}{RGB}{0,153,255}
\definecolor{llava_red}{RGB}{204,0,0}
\definecolor{gemma_blue}{RGB}{0,128,255}
\definecolor{deepseek_teal}{RGB}{0,153,136}
\definecolor{mydarkblue}{rgb}{0,0.53,0.96}
\definecolor{backcolor}{RGB}{245,248,250}
\definecolor{emph}{RGB}{166,88,53}
\definecolor{nightblue}{RGB}{9,49,105}
\definecolor{keywords}{RGB}{207,33,46}
\definecolor{lightpurple}{RGB}{130,81,223}
\definecolor{CLIPBlue}{rgb}{0.192,0.454,0.643}
\hypersetup{
  citecolor=mydarkblue,
  linkcolor=mydarkblue,
  urlcolor=mydarkblue
}

\definecolor{darkorange}{RGB}{204,102,0}
\definecolor{editpink}{RGB}{219,48,122}
\definecolor{editblue}{RGB}{0,100,200}
\definecolor{editpurple}{RGB}{112,48,160}

\definecolor{editgreen}{RGB}{0,140,60}
\newcommand{\opusfour}{\model{Claude~Opus~\textcolor{claude_orange}{4.6}}\xspace}

\newcommand{\gptfivefour}{\model{Codex-\textcolor{gpt_green}{5.4}}\xspace}
\newcommand{\opusshort}{\model{Opus-\textcolor{claude_orange}{4.6}}\xspace}
\newcommand{\gptshort}{\model{Codex-\textcolor{gpt_green}{5.4}}\xspace}
\newcommand{\geminipronew}{\model{Gemini~Pro~\textcolor{gemini_blue}{3.1}}\xspace}

\newcommand{\datasetname}{\texttt{AutoWorldModel-Bench}\xspace}

\newcommand{\asteroids}{\textsc{Asteroids}\xspace}
\newcommand{\breakout}{\textsc{Breakout}\xspace}
\newcommand{\frogger}{\textsc{Frogger}\xspace}
\newcommand{\kong}{\textsc{Kong}\xspace}
\newcommand{\platformer}{\textsc{Platformer}\xspace}
\newcommand{\pong}{\textsc{Pong}\xspace}
\newcommand{\racer}{\textsc{Racer}\xspace}
\newcommand{\snake}{\textsc{Snake}\xspace}

\newlength{\legendspace}
\newcolumntype{C}[1]{>{\centering\arraybackslash}m{#1}}

\newcommand{\papertitle}[0]{\texttt{AutoWorldModel-Bench}: A State-Centric Benchmark for Automated World-Model Research}
\title{\papertitle}

\author{%
  Marjan Moodi \\
  Electronic Arts \\
  \texttt{mmoodi@ea.com} \\
  \And
  Xuankang Zhu\thanks{Work done during an internship at EA.} \\
  Simon Fraser University \\
  \texttt{jacobz0106@gmail.com} \\
  \And
  Fernando De Mesentier Silva \\
  Electronic Arts \\
  \texttt{fdemesentiersilva@ea.com} \\
  \AND
  Harold Chaput \\
  Electronic Arts \\
  \texttt{hchaput@ea.com} \\
  \And
  Mohammad Reza Taesiri \\
  Electronic Arts \\
  \texttt{mtaesiri@ea.com} \\
}

\lstdefinestyle{mystyle}{
    backgroundcolor=\color{backcolor},
    commentstyle=\color{ForestGreen},
    keywordstyle=\color{keywords},
    stringstyle=\color{nightblue},
    basicstyle=\ttfamily\footnotesize,
    breakatwhitespace=false,
    breaklines=true,
    captionpos=b,
    keepspaces=true,
    showspaces=false,
    showstringspaces=false,
    showtabs=false,
    tabsize=2,
    frame=shadowbox,
    emph={AutoTokenizer,AutoModelForSequenceClassification,Explainer},
    emphstyle={\color{emph}},
    emph={[2]from_pretrained,compute_table},
    emphstyle={[2]\color{lightpurple}}
}
\newcommand{\subsec}[1]{\noindent\textbf{#1}~~}

\crefname{figure}{Fig.}{Figs.}
\Crefname{figure}{Fig.}{Figs.}
\crefname{table}{Tab.}{Tabs.}
\Crefname{table}{Tab.}{Tabs.}
\crefname{section}{Sec.}{Secs.}
\Crefname{section}{Sec.}{Secs.}

\makeatletter
\newcommand{\appendixtitleblock}{%
  \thispagestyle{empty}%
  \vbox{%
    \hsize\textwidth
    \linewidth\hsize
    \vskip 0.1in
    \@toptitlebar
    \centering
    {\LARGE\bf Appendix for: \@title\par}
    \@bottomtitlebar
    \vskip 0.25in
  }%
}
\makeatother

\begin{document}

\maketitle

\begin{abstract}
World modeling is an unsettled field: architectures, training objectives, and state representations interact in complex ways, and no single recipe dominates across environments. This makes it an ideal testbed for AI coding agents acting as autonomous researchers---a setting in which the improvement direction is not specified in advance, unlike the engineering-to-spec tasks that dominate current agent benchmarks.
We introduce \datasetname{}, a closed-loop benchmark in which frontier coding agents autonomously improve a provided base world model under a fixed compute budget.
The benchmark spans eight game environments under a unified structured-state representation---ground-truth entity state extracted from each game and consumed through a shared tensor format---which isolates dynamics modeling from perception and enables minutes-per-run iteration.
Across $64$ sessions, \gptfivefour and \opusfour improve their base on a held-out test split in all but one session, with about half ($33$ of $64$) a substantial gain ($\Delta\ge+0.10$) and the remaining improvements smaller but positive; in $91\%$ of sessions the winning edit is a substantive change to the model or training rather than a hyperparameter tweak. Our benchmark offers a setting in which frontier coding agents can be evaluated on open-ended research rather than engineering-to-spec problems.
{Project page: \url{https://electronicarts.github.io/AutoWorldModelBench/}.}
\end{abstract}

\section{Introduction}
\label{sec:introduction}

Learning predictive models of environment dynamics---referred to as world modeling---has become one of the most active frontiers in artificial intelligence~\citep{ha2018world,lecun2022path}.
In just the past few years, learned world models have progressed from simple latent dynamics predictors to systems that can imagine long-horizon trajectories, plan in learned state spaces, and even simulate entire interactive video games~\citep{hafner2023mastering,schrittwieser2020muzero,bruce2024genie,valevski2024gamengen}.
This rapid progress is driven by a rich and growing design space: researchers must choose among fundamentally different state representations, dynamics architectures, training objectives, and rollout strategies, with each combination yielding different tradeoffs in accuracy, generalization, and computational cost~\citep{hafner2019planet,micheli2023transformers,hansen2024tdmpc2,alonso2024diffusion}.
No single approach dominates, and the field continues to evolve quickly---making world modeling both a scientifically important problem and one where systematic exploration of the design space could yield substantial returns.

At the same time, a new paradigm is emerging in how research itself is conducted.
Autonomous AI research agents---systems capable of reading codebases, formulating hypotheses, writing training pipelines, running experiments, and iterating on results---have demonstrated that meaningful scientific contributions can be produced with minimal human intervention~\citep{lu2024aiscientist,schmidgall2025robin,su2025mlzero,liu2025mlagent}.
The potential of such systems to accelerate progress is immense, but so is the need to evaluate them rigorously.
Current evaluations of AI research agents span a growing spectrum, from machine-learning engineering tasks with well-defined objectives and external metrics~\citep{chan2024mlebench} to broader ML research and replication~\citep{nathani2025mlgym,nie2025mlrbench,staab2025paperbench} and early scientific-discovery settings~\citep{chen2024scienceagentbench,jansen2024discoveryworld}.
Yet the landscape remains uneven: most benchmarks are concentrated on well-scoped problems where datasets, metrics, and improvement directions are largely specified in advance, while the most open-ended settings are harder to evaluate in a controlled and reproducible way.
What remains scarce are evaluations that combine both properties---placing agents at genuine research frontiers where the design space is vast, interactions among decisions are poorly understood, experiments unfold sequentially under limited budgets, and success depends on generating and testing novel hypotheses rather than recombining known recipes.

In this paper, we bring these two trajectories together.
We introduce a benchmark for world-model research built on a unified structured-state representation across diverse game environments, together with pipelines for data extraction, preparation, and model evaluation.
We make the following contributions:
\begin{compactenum}
    \item We introduce \datasetname{}, a benchmark over eight game environments with a unified structured-state representation that sidesteps perception and supports minutes-per-run iteration (\Cref{sec:method}).
    \item We define a six-hour compute-bounded closed-loop protocol in which an agent receives a base world model architecture and must autonomously improve it (\Cref{sec:harness}).
    \item We evaluate \gptfivefour and  \opusfour across the $64$ (game, base) sessions and analyze where the improvements come from: agents lift the base on $63$ of $64$ sessions, and the gain is concentrated at long-horizon rollout rather than one-step fit (\Cref{sec:results-reliable-improvement,sec:results-horizon}).
    \item We classify every agent experiment against its base using a zero-shot judge and find that in $91\%$ of sessions the winning edit is a substantive, non-trivial change to the model or training rather than a scalar-knob adjustment (\Cref{sec:results-nontrivial}).
\end{compactenum}

\section{Related Work}
\label{sec:related}

\subsec{World models.}
World models learn action-conditioned dynamics for planning, control, and simulation. Prior work ranges from early action-conditioned video prediction and latent dynamics models for model-based reinforcement learning~\citep{oh2015action,ha2018world,hafner2019planet,hafner2020dream,hafner2021mastering,hafner2023mastering,schrittwieser2020muzero} to visually grounded game simulators and more recent large-scale generative models~\citep{simple2019,micheli2023transformers,twm2023,storm2023,alonso2024diffusion,bruce2024genie,valevski2024gamengen,jowa2024}. Much of this literature learns from pixels or latent variables inferred from pixels, so perceptual and dynamics errors are intertwined. A complementary state-based line learns dynamics over compact state for downstream control~\citep{chua2018pets,janner2019mbpo,hansen2024tdmpc2}. We adopt this state-centric perspective, but provide exact simulator state to isolate transition modeling itself and enable fast iteration for automated world-model research.

\subsec{Object-centric and entity-centric representations.}
A long line of work argues that factored representations make dynamics more compositional. In perception, object-centric methods infer sets of objects or slots from images and video~\citep{eslami2016attend,kosiorek2018sqair,burgess2019monet,greff2019iodine,locatello2020slot,kipf2022savi,elsayed2022savipp,singh2022steve}. In dynamics, relational and object-centric models predict transitions over objects, interactions, or slots rather than monolithic frame embeddings~\citep{battaglia2016interaction,watters2017visual,kipf2018neural,zhu2018oodp,veerapaneni2019op3,kipf2020contrastive,wu2022slotformer,wu2024sswm}. These approaches typically must infer object identity and correspondence, so perception errors can propagate into the learned dynamics. In contrast, video games already maintain structured entity state internally; our benchmark exposes that structured state directly, enabling controlled study of world-model learning over grounded entities.

\subsec{Automated AI research agents.}
Recent LLM-based systems have begun to automate substantial parts of the ML research loop, including code synthesis, experiment design, execution, analysis, and iterative refinement~\citep{lu2024aiscientist,li2024mlrcopilot,schmidgall2025robin,su2025mlzero,liu2025mlagent}. Evaluation suites have expanded in parallel, covering ML engineering, open-ended research workflows, replication, and scientific-discovery settings~\citep{chan2024mlebench,nathani2025mlgym,rebench2024,staab2025paperbench,nie2025mlrbench,chen2024scienceagentbench,jansen2024discoveryworld}. Our benchmark targets a narrower but scientifically active setting: agents must improve world models under a fixed compute budget in a large, underdetermined design space. This preserves the iterative character of research while keeping success measurable through held-out predictive performance of the discovered model.

\subsec{World-model evaluation.}
Evaluating world models requires more than one-step prediction error. Prior work has emphasized complementary desiderata such as interactive physical reasoning and action fidelity~\citep{bakhtin2019phyre,actbench2024}. {Recent benchmarks evaluate world-model learning beyond next-frame prediction, including WorldTest/AutumnBench~\citep{benchmarkingworldmodellearning2025} for reward-free exploration and derived test tasks, and WorldModelBench~\citep{worldmodelbench2025} which judges video-generation models as world models. Our benchmark differs from these in two ways: (i) it is state-centric rather than pixel- or video-based, and (ii) it is closed-loop on the agent side---the model under test is produced by an autonomous coding agent within a fixed compute budget, rather than being a pre-trained model under evaluation.} 
Our protocol targets state-centric world models: since \datasetname{} provides exact structured-state trajectories, we evaluate action-conditioned prediction at the entity-state level under teacher-forced and multi-step open-loop rollouts. We also include a held-out scenario suite from grounded initial states that probes mechanics such as collisions, scoring events, and terminal conditions, testing models on both average trajectory fit and rule consistency/action responsiveness.

\section{\datasetname{}}
\label{sec:method}

\subsection{State Representation}
\label{sec:ecs}

\begin{figure}[t]
\centering
\begin{subfigure}[c]{0.35\linewidth}
  \centering
  \includegraphics[width=\linewidth]{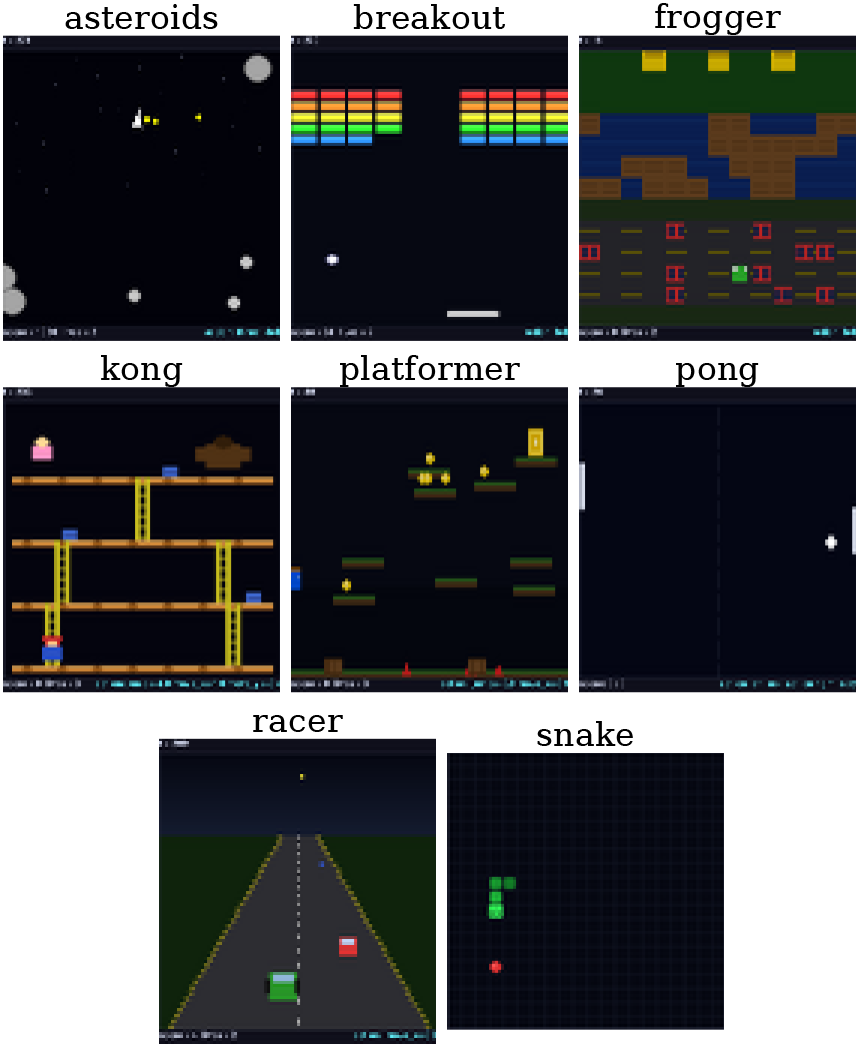}
  \caption{}
  \label{fig:game-samples}
\end{subfigure}
\hfill
\begin{subfigure}[c]{0.63\linewidth}
  \centering
  \includegraphics[width=\linewidth]{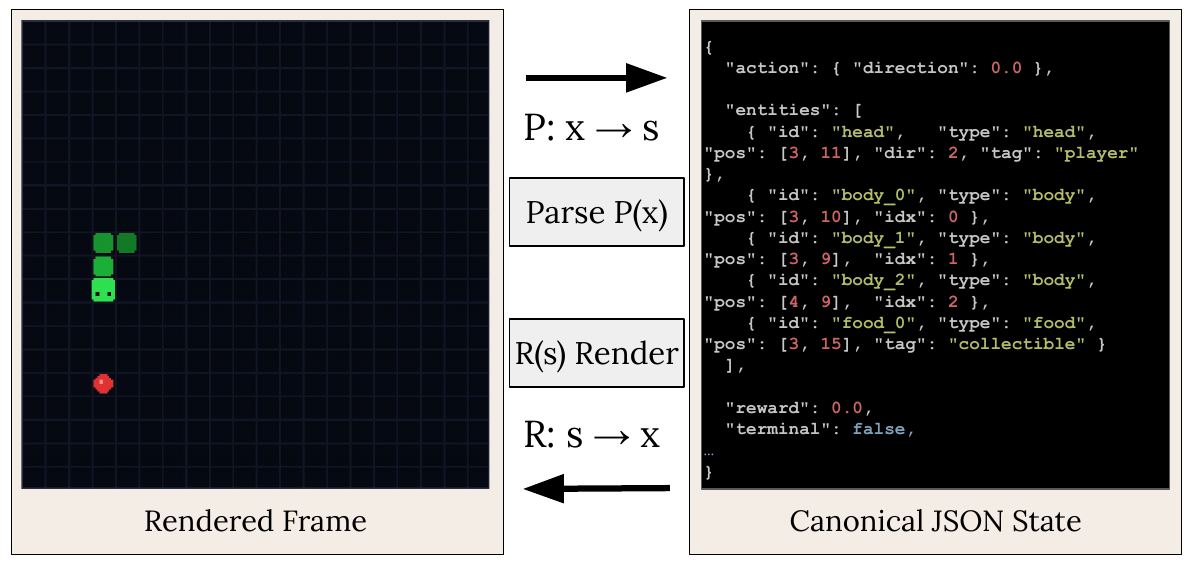}
  \caption{}
  \label{fig:ecs-render}
\end{subfigure}
\caption{\textbf{(a)}~The eight benchmark games span classic arcade genres with varying entity counts, dynamics (grid, continuous physics, multi-agent), and termination conditions. \textbf{(b)} Each game frame pairs a visual render with the simulator's ground-truth structured entity state (shown for \snake), enabling world modeling directly on ground-truth entity state.}
\label{fig:ecs-pipeline}
\end{figure}

In \datasetname{} every game state is represented as an \emph{entity-component-system} (ECS) snapshot (\Cref{fig:ecs-pipeline}).
ECS is a well-established architectural pattern in game engines~\cite{redmond2025ecs, bilas2002,west2007,martin2007a,west2018} that decomposes a scene into three orthogonal elements:
\textbf{entities} are uniquely identified objects in the world (a player character, a projectile, a wall);
\textbf{components} are typed data containers attached to entities (position, velocity, collision shape, visual appearance);
and \textbf{systems} are functions that operate over entities sharing particular component signatures (a physics system updates all entities with a transform and a physics component).
This decomposition separates \emph{what exists} from its \emph{properties} and \emph{behavior}, yielding modular, compositional, and easily serialized representations.

Each game frame is serialized as a \emph{frame envelope}---a JSON object recording the player's action, global state (score, lives, game-specific counters), and an ordered list of entity slots.
Each slot holds a kind label (e.g., \texttt{head}, \texttt{ball}), an alive flag, and a subset of five typed components: \textbf{Transform} (position, rotation, scale), \textbf{Physics} (velocity, mass, restitution, damping), \textbf{Collider} (shape, extents, collision layer), \textbf{Material} (color, material ID), and \textbf{Gameplay} (hit points, semantic flags, game-specific stats).
Components attach optionally, so the same schema accommodates games with very different internal structures---a frame from \snake (grid-snapped positions, discrete actions) and a frame from \asteroids (continuous positions, thrust and rotation) share identical structure.
Full field-level specifications and a worked example are provided in \Cref{sec:appendix-envelope}.

\begin{table}[t]
\centering
\caption{Benchmark environments and statistics. \emph{Ent.}: max entity slots per frame. \emph{Act.}: semantic player controls, with type in parentheses (discrete = class index; continuous = scalar). \emph{\#F}: active fields in the unified $7$-dim action vector; other slots are masked. \emph{Policy}: episode-collection agents (H = heuristic, R = random, RL = reinforcement learning).
}
\label{tab:games}
\small
\setlength{\tabcolsep}{2.5pt}
\resizebox{\textwidth}{!}{%
\begin{tabular}{lclccrrrl}
\toprule
\textbf{Game} & \textbf{Ent.} & \textbf{Act.} & \textbf{\#F} & \textbf{Policy} & \textbf{Ep.} & \textbf{Frames} & \textbf{Avg Len} & \textbf{Key Challenge} \\
\midrule
Snake          & $\leq$48  & direction (discrete, 4 classes)                    & 1 & RL/H/R & 19k & 15.9M &    837 & Growth, self-collision \\
Frogger        & $\leq$28  & action (discrete, 5 classes)                       & 1 & RL/H/R & 19k &  6.25M &   329 & Lane traffic, navigation \\
Pong           &  5        & paddle\_l\_dy, paddle\_r\_dy (continuous)          & 2 & RL/H/R & 19k & 42.5M & 2{,}237 & Ball--paddle physics \\
Breakout       & $\leq$52  & action (discrete, 3 classes)                       & 1 & H/R    & 19k & 33.0M & 1{,}739 & Brick destruction \\
Asteroids      & $\leq$20  & action (discrete, 8 classes)                       & 1 & H/R    & 19k &  5.64M &   297 & Splitting, wrapping \\
Platformer     & $\leq$24  & jump, move\_x (continuous)                         & 2 & RL/H/R & 19k & 10.6M &    556 & Gravity, collision \\
Kong           & $\leq$16  & jump, move\_x, move\_y (continuous)                & 3 & RL/H/R & 19k & 23.0M & 1{,}209 & Ladders, barrels \\
Racer          &  6        & move\_x (continuous)                               & 1 & RL/H/R & 19k & 21.1M & 1{,}112 & Lane switching, speed \\
\midrule
\textbf{Total} &           &                                          &                          &        & \textbf{152k} & \textbf{158.0M} & & \\
\bottomrule
\end{tabular}%
}
\end{table}

Our benchmark includes eight game environments covering diverse dynamics and challenges (\Cref{tab:games}).
Each uses a \emph{game adapter} to convert engine state into the canonical frame envelope, paired with a \emph{manifest} declaring entity kinds, mutability, action fields, and coordinate bounds.

\subsection{Tensor Representation}
\label{sec:tensor}

Each game defines a fixed slot budget $N$ equal to the maximum entity count observed across all training episodes (e.g., $N{=}5$ for \pong: two paddles, ball, and two boundary walls at top and bottom).
Every episode uses the same $N$ slots; unused positions are marked \emph{dead} via an entity field.
A per-game manifest declares each entity kind as either \emph{mutable} (prediction target: snake head, ball, asteroid) or \emph{immutable} (conditioning context: boundary walls, static platforms, goal tiles).
All dimensions are unified across games---unused fields are zeroed via per-game masks, so models see identically shaped tensors regardless of game.
Each episode is represented as follows (\Cref{sec:appendix-formatd}):

\begin{compactitem}
    \item \textbf{Entity registry} $\mathbf{R} \in \mathbb{R}^{N \times 34}$: static physical identity of each entity---collider shape (one-hot, 4), radius (1), half-extents (2), collision layer (one-hot, 18), trigger flag (1), mutability flag (1), scale (2), and physics material (5).  Constant within an episode.
    \item \textbf{Entity state} $\mathbf{S}_t \in \mathbb{R}^{N \times 23}$: per-frame dynamics of each entity---normalized position (2), alive flag (1), velocity (2), gameplay fields (14, per-game masked), and position history deltas (4).
    \item \textbf{Player action} $\mathbf{a}_t \in \mathbb{R}^{7}$: the control input at frame $t$. The 7 dimensions are the union of per-game action fields (\texttt{action}, \texttt{direction}, \texttt{jump}, \texttt{move\_x}, \texttt{move\_y}, \texttt{paddle\_l\_dy}, \texttt{paddle\_r\_dy}); each game activates 1--3 via a per-game mask (\Cref{tab:games}). This vector flat-concatenates semantic fields rather than forming a product of discrete choices: a discrete field stores one class index (e.g., Snake's \texttt{direction} $\in\{0,\ldots,3\}$ is one slot, not four), while a continuous field stores one scalar (e.g., Pong's \texttt{paddle\_l\_dy}).
    \item \textbf{Game state} $\mathbf{g}_t \in \mathbb{R}^{17}$: non-entity game-level variables observed at frame $t$---scores, remaining lives, level counters, and similar bookkeeping that is not tied to any individual entity.  Unified to 17 dimensions; per-game masked.
    \item \textbf{Terminal flag} $t_t \in \{0,1\}$: indicates whether the episode ends at frame $t$.
\end{compactitem}

\subsection{Data Collection}
\label{sec:datagen}

The dataset contains $152{,}000$ episodes totaling over $158$ million frames across the eight benchmark games (\Cref{tab:games}).
Episodes are collected using three play policies: game-specific heuristic agents, random agents, and RL agents trained with PPO~\citep{schulman2017ppo} or DQN~\citep{mnih2015dqn} with per-game reward shaping (\Cref{sec:appendix-reward-shaping}).
After manual review of the collected trajectories, we added RL agents for six games where heuristic play alone failed to reliably reach late-game states or produce completed episodes.

\subsection{World-Model Evaluation Protocol}
\label{sec:eval}

We evaluate learned world models through three complementary modes:
\begin{compactitem}
    \item \textbf{Teacher-forced ($h{=}1$):} The model receives ground-truth $s_t$ and predicts $\hat{s}_{t+1}$, isolating single-step dynamics accuracy from error accumulation.
    \item \textbf{Open-loop rollout ($h{>}1$):} Given initial state $s_0$ and a recorded action sequence, the model autoregressively predicts $(\hat{s}_1, \ldots, \hat{s}_T)$ by feeding its own outputs back, measuring error compounding over horizons $h \in \{10, 20\}$. The full input/output contract is in \Cref{sec:appendix-contract}.
    \item \textbf{Scenario tests:} Per-game probes of whether the model learns game rules, not just trajectory statistics. Each scenario builds a controlled initial state isolating one mechanic (e.g., ball--paddle collision, food consumption), applies a deterministic action sequence, and checks predictions against ground truth. Details are in~\Cref{sec:appendix-scenarios}.
\end{compactitem}

\begin{figure}[t]
  \centering
  \includegraphics[width=\linewidth]{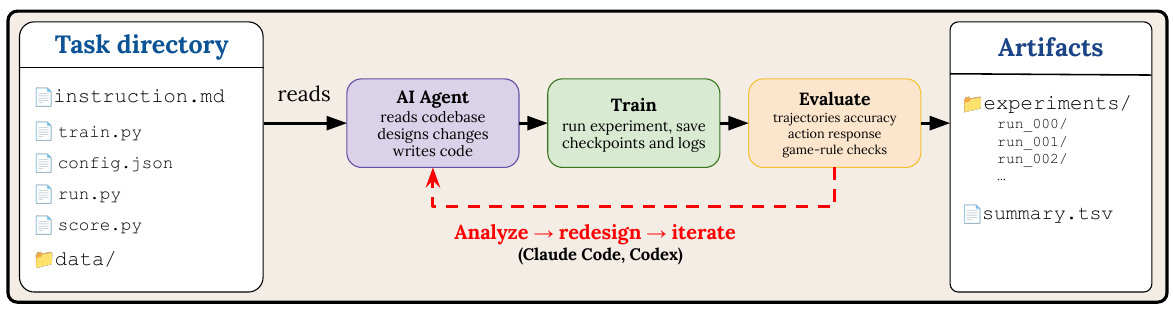}
  \caption{Our closed-loop agent harness. Each task provides a self-contained directory with an \texttt{instruction.md}, a base model, training data, and evaluation scripts. The AI agent reads the instructions, designs improvements, trains a model (capped at 10 minutes per run), and evaluates it. All experiments are logged in a structured \texttt{experiments/} directory with a \texttt{summary.tsv} index.}
  \label{fig:agent-loop}
\end{figure}

\subsection{Agent Harness}
\label{sec:harness}

We adopt a closed-loop design in which each task is presented with a self-contained directory with standardized preparation and training scripts~\citep{karpathy2025autoresearch}. Following~\citet{lee2026metaharness}, we give the agent persistent access to a structured experiment log that records the full history of prior runs, including architectures explored, hyperparameters used, and scores obtained. This setup allows each subsequent decision to be conditioned on all accumulated evidence (\Cref{fig:agent-loop}).

\subsec{Task structure.}
Each task directory provides everything an agent needs:
\begin{compactitem}
    \item \textbf{Instructions} (\texttt{instruction.md}): research goal, scoring formula, file permissions, data format, constraints, and tracking conventions. The same template is used for every (game, base) pair with only \{game\}, \{model\}, and entity count substituted (\Cref{sec:appendix-instruction}).
    \item \textbf{Base world model} (\texttt{train.py}): a single-file world model---architecture, loss, and training loop---serving as the agent's starting point. The agent may freely modify this file.
    \item \textbf{Configuration} (\texttt{config.json}): hyperparameters (learning rate, batch size, window size, training budget). The agent creates configs under \texttt{configs/} for each experiment.
    \item \textbf{Experiment runner} (\texttt{run.py}): wraps \texttt{train.py} with tracking---assigns sequential run IDs, copies outputs to \texttt{experiments/<run\_id>/}, and appends results to \texttt{summary.tsv}. Read-only.
    \item \textbf{Scorer} (\texttt{score.py}): independently evaluates a checkpoint on the validation set at all three horizons. Read-only.
    \item \textbf{Data}: pre-cached windowed tensors ($W{=}8$) for training and validation, mounted read-only.
\end{compactitem}

\subsec{Agent loop.}
The agent studies the base code and iterates: \emph{analyze} \texttt{summary.tsv} and prior logs, \emph{design} a change (architecture, loss, or hyperparameter), \emph{train} it via \texttt{run.py}, and \emph{evaluate} it.
At session end, an independent verifier scores the best checkpoint via \texttt{score.py} and reports the weighted final score.

\subsec{Execution infrastructure.}
The agent harness is implemented on top of Harbor~\citep{harbor2025}, a containerized task runtime for AI agents. Each task is packaged as a Docker container with the pre-cached training data mounted read-only and instructions declaring the game, model, and compute budget.

\subsec{Safeguards against reward hacking.} Because the agent may freely modify \texttt{train.py}, the harness is designed so that a model cannot gain by optimizing the scorer rather than the dynamics. \texttt{run.py}, \texttt{score.py}, and the evaluator are mounted read-only, and the scorer hard-defaults its evaluation horizons to $\{1,10,20\}$, so config edits cannot narrow what is scored. The scored deliverable is a trained checkpoint re-evaluated \emph{after} the session by an independent evaluator on the test and scenario splits, which the agent never sees, so an in-session score cannot be carried over. The Alive-F1 term in the composite defeats degenerate ``predict everything alive'' solutions.

\section{Experiments}
\label{sec:experiments}
\label{sec:exp-setup}

We use \datasetname{} to benchmark frontier AI coding agents as automated world-model researchers, measuring how effectively they can improve provided base models.

\subsec{Base models.}
We design four bases covering different architecture families:
\begin{compactitem}
    \item \textbf{RSSM/Dreamer}~\citep{hafner2023mastering} --- GRU-based recurrent state-space model with discrete categorical latent ($32{\times}32$), free-bits KL regularization, and symlog loss.
    \item \textbf{AR-Transformer}~\citep{micheli2023transformers} --- causal autoregressive Transformer with block-causal attention. Unlike IRIS, which operates on discrete image tokens, our variant works directly on continuous structured entity state.
    \item \textbf{D3PM}~\citep{austin2021d3pm} --- Transformer encoder with discrete denoising diffusion, quantizing positions and gameplay fields into per-field token vocabularies.
    \item \textbf{MaskGIT}~\citep{chang2022maskgit} --- Transformer encoder with masked-generation objective and iterative parallel decoding.
\end{compactitem}
All four operate on the tensorized structured-state representation (\Cref{sec:tensor}) and predict next-frame entity positions, alive status, terminal flag, and per-game gameplay fields.
The continuous models (Dreamer, AR-Transformer) predict position deltas relative to the current state; the discrete models (D3PM, MaskGIT) quantize positions into per-axis token vocabularies and predict absolute bin indices.
All share a velocity consistency loss ($\lambda{=}0.1$).
AR-Transformer, D3PM, and MaskGIT use the same Transformer backbone (block-causal temporal context encoder); Dreamer uses GRU-based recurrence with mean-pooling over entities.
Full architecture details are provided in \Cref{sec:appendix-models}.

\subsec{Agents and compute budget.}
We evaluate two frontier coding agents, \opusfour and \gptfivefour, each under an identical budget: a single H100 GPU for a $6$-hour session, with a $10$-minute wall-clock cap per individual training run. Every (game, base) pair is evaluated exactly once per agent, producing $64$ sessions in total ($2$ agents $\times$ $8$ games $\times$ $4$ bases).

\subsec{Data splits.}
Each game's trajectories are partitioned into \emph{training}, \emph{validation}, \emph{test}, and \emph{scenario} splits. The agent has read-only access to the training and validation splits during its session; the test and scenario splits are held out and used only for the final post-session evaluation.

\subsec{Per-horizon composite.}
At each rollout horizon $h$ we measure two quantities: \emph{Position~L1}, the mean $L_1$ error on mutable entity positions (normalized to $[0,1]$; $\downarrow$), and \emph{Alive~F1}, the $F_1$ score on the entity alive/dead flag ($\uparrow$). These combine into a per-horizon composite
\begin{equation*}
\text{composite}_h \;=\; 0.9 \cdot \bigl(1 - \text{Position\,L1}_h\bigr) \;+\; 0.1 \cdot \text{Alive\,F1}_h.
\end{equation*}
The weighting reflects that Position~L1 carries the dominant dynamical signal while Alive~F1 guards against trivial solutions that predict every entity as alive.

\subsec{Final score.}
We rank models by a horizon-weighted score favoring long-horizon rollout accuracy:
\begin{equation*}
\text{final} \;=\; 0.1 \cdot \text{composite}_1 \;+\; 0.2 \cdot \text{composite}_{10} \;+\; 0.7 \cdot \text{composite}_{20}.
\end{equation*}
During each session, the agent sees this validation score to guide search. Afterward, we recompute it on two held-out splits: the \emph{test} split, giving the \emph{test score}, and the \emph{scenario suite}, giving the \emph{scenario score}. Both use the same $0.9 \cdot (1 - \text{PositionL1}) + 0.1 \cdot \text{AliveF1}$ per-horizon composite. The test score blends horizons $h_1, h_{10}, h_{20}$ with weights $0.1, 0.2, 0.7$. The scenario score uses the same horizon weighting but evaluates the long-horizon term at the \emph{end of episode} ($h_{\mathrm{end}}$) rather than at $h_{20}$: $\text{scenario} = 0.1 \cdot c_1 + 0.2 \cdot c_{10} + 0.7 \cdot c_{h_{\mathrm{end}}}$. This rewards agents whose models remain faithful over the full curated rollout rather than only at a truncated $20$-step horizon. Per-horizon decomposition and additional diagnostics (e.g., the terminal-correct signal) are reported in \Cref{sec:supp-scenarios}.
We select the best checkpoint by validation performance---the session's \emph{winning edit}---then evaluate it once on the held-out test split and once on the scenario suite. A session \emph{improves the base} when this winning edit raises the held-out composite above the base's.
The test score is the primary metric in \Cref{sec:results}; the scenario score is a robustness check.

\subsec{Search baselines.}
To test whether tuning the exposed hyperparameters alone reaches the agents' gains, we run two compute-matched search baselines in every cell: random search and Bayesian optimization (Optuna/TPE~\citep{akiba2019optuna,bergstra2012random}). Each search optimizes the same validation composite through the identical \texttt{run.py} and \texttt{score.py}, under the agents' exact per-cell budget of a $6$-hour session with a $10$-minute per-run cap, and is scored on the same held-out test and scenario splits.

\subsec{Seed repetition.}
Because each session is run once, we repeat the full session under two additional random seeds on half the grid (\asteroids, \snake, \frogger, and \racer, all four bases, both agents), giving three independent sessions per cell and $96$ in total. Each repetition restarts the agent from the base, so it reruns the entire discovery process rather than re-scoring a fixed checkpoint.

\section{Results}
\label{sec:results}

\begin{table}[t]
\centering
\small
\caption{Per-(game, base) lift on two held-out evaluations. ``Base'' is the base test score. ``$\Delta$ test'' is the session-best improvement over the base on the held-out test split, using the horizon weights $0.1 \cdot c_1 + 0.2 \cdot c_{10} + 0.7 \cdot c_{20}$. ``$\Delta$ scenario'' uses the same composite per horizon, but blends $0.1 \cdot c_1 + 0.2 \cdot c_{10} + 0.7 \cdot c_{h_{\mathrm{end}}}$ so the long-horizon term reflects the full curated rollout length instead of a truncated $20$ steps. Each $\Delta$ column uses a divergent palette: $\Delta$ test uses blue/crimson for gains/regressions (saturated at $|\Delta|{=}0.70$), $\Delta$ scenario uses amber/violet (saturated at $|\Delta|{=}0.60$).}
\label{tab:cell-matrix}
\setlength{\tabcolsep}{4pt}
\resizebox{\textwidth}{!}{%
  \begin{tabular}{l l cc cc cc cc}
\toprule
 & & \multicolumn{2}{c}{Base (test)} & \multicolumn{2}{c}{Best (test)} & \multicolumn{2}{c}{$\Delta$ test} & \multicolumn{2}{c}{$\Delta$ scenario} \\
\cmidrule(lr){3-4}\cmidrule(lr){5-6}\cmidrule(lr){7-8}\cmidrule(lr){9-10}
Game & Arch. & \opusshort & \gptshort & \opusshort & \gptshort & \opusshort & \gptshort & \opusshort & \gptshort \\
\midrule
 \multirow{4}{*}{\asteroids} & Dreamer & 0.09 & 0.09 & 0.69 & 0.71 & \cellcolor{tabblue!86}+0.60 & \cellcolor{tabblue!87}+0.61 & \cellcolor{tabamber!70}+0.42 & \cellcolor{tabamber!71}+0.43 \\
  & AR-Trans. & 0.10 & 0.10 & 0.70 & 0.68 & \cellcolor{tabblue!86}+0.60 & \cellcolor{tabblue!84}+0.59 & \cellcolor{tabamber!78}+0.47 & \cellcolor{tabamber!72}+0.43 \\
  & D3PM & 0.32 & 0.30 & 0.52 & 0.64 & \cellcolor{tabblue!29}+0.20 & \cellcolor{tabblue!48}+0.34 & \cellcolor{tabamber!35}+0.21 & \cellcolor{tabamber!51}+0.31 \\
  & MaskGIT & 0.38 & 0.33 & 0.55 & 0.46 & \cellcolor{tabblue!23}+0.16 & \cellcolor{tabblue!19}+0.13 & \cellcolor{tabamber!33}+0.20 & \cellcolor{tabamber!26}+0.16 \\
\cmidrule(lr){2-10}
 \multirow{4}{*}{\breakout} & Dreamer & 0.94 & 0.94 & 0.97 & 0.97 & \cellcolor{tabblue!4}+0.03 & \cellcolor{tabblue!4}+0.03 & \cellcolor{tabamber!15}+0.09 & \cellcolor{tabamber!13}+0.08 \\
  & AR-Trans. & 0.73 & 0.69 & 0.99 & 0.98 & \cellcolor{tabblue!37}+0.26 & \cellcolor{tabblue!41}+0.29 & \cellcolor{tabamber!74}+0.44 & \cellcolor{tabamber!52}+0.31 \\
  & D3PM & 0.66 & 0.65 & 0.66 & 0.68 & \cellcolor{tabred!30}-0.00 & \cellcolor{tabblue!4}+0.03 & \cellcolor{tabamber!1}+0.00 & \cellcolor{tabamber!5}+0.03 \\
  & MaskGIT & 0.59 & 0.62 & 0.66 & 0.98 & \cellcolor{tabblue!10}+0.07 & \cellcolor{tabblue!52}+0.36 & \cellcolor{tabamber!14}+0.09 & \cellcolor{tabamber!52}+0.31 \\
\cmidrule(lr){2-10}
 \multirow{4}{*}{\frogger} & Dreamer & 0.69 & 0.71 & 0.74 & 0.78 & \cellcolor{tabblue!7}+0.05 & \cellcolor{tabblue!10}+0.07 & \cellcolor{tabamber!24}+0.14 & \cellcolor{tabamber!1}+0.01 \\
  & AR-Trans. & 0.24 & 0.41 & 0.72 & 0.75 & \cellcolor{tabblue!68}+0.48 & \cellcolor{tabblue!49}+0.34 & \cellcolor{tabamber!73}+0.44 & \cellcolor{tabamber!48}+0.29 \\
  & D3PM & 0.65 & 0.64 & 0.72 & 0.67 & \cellcolor{tabblue!10}+0.07 & \cellcolor{tabblue!5}+0.03 & \cellcolor{tabamber!12}+0.07 & \cellcolor{tabamber!20}+0.12 \\
  & MaskGIT & 0.67 & 0.66 & 0.73 & 0.93 & \cellcolor{tabblue!10}+0.07 & \cellcolor{tabblue!40}+0.28 & \cellcolor{tabamber!11}+0.06 & \cellcolor{tabamber!43}+0.26 \\
\cmidrule(lr){2-10}
 \multirow{4}{*}{\kong} & Dreamer & 0.50 & 0.78 & 0.83 & 0.84 & \cellcolor{tabblue!47}+0.33 & \cellcolor{tabblue!9}+0.06 & \cellcolor{tabamber!66}+0.40 & \cellcolor{tabamber!38}+0.23 \\
  & AR-Trans. & 0.59 & 0.37 & 0.78 & 0.84 & \cellcolor{tabblue!26}+0.18 & \cellcolor{tabblue!68}+0.48 & \cellcolor{tabamber!35}+0.21 & \cellcolor{tabamber!53}+0.32 \\
  & D3PM & 0.47 & 0.49 & 0.61 & 0.96 & \cellcolor{tabblue!20}+0.14 & \cellcolor{tabblue!67}+0.47 & \cellcolor{tabamber!26}+0.16 & \cellcolor{tabamber!48}+0.29 \\
  & MaskGIT & 0.50 & 0.48 & 0.57 & 0.61 & \cellcolor{tabblue!10}+0.07 & \cellcolor{tabblue!18}+0.12 & \cellcolor{tabamber!17}+0.10 & \cellcolor{tabamber!23}+0.14 \\
\cmidrule(lr){2-10}
 \multirow{4}{*}{\platformer} & Dreamer & 0.95 & 0.92 & 0.96 & 0.96 & \cellcolor{tabblue!1}+0.01 & \cellcolor{tabblue!6}+0.04 & \cellcolor{tabamber!1}+0.01 & \cellcolor{tabamber!14}+0.08 \\
  & AR-Trans. & 0.72 & 0.73 & 0.96 & 0.85 & \cellcolor{tabblue!34}+0.24 & \cellcolor{tabblue!17}+0.12 & \cellcolor{tabamber!67}+0.40 & \cellcolor{tabamber!29}+0.18 \\
  & D3PM & 0.62 & 0.64 & 0.97 & 0.97 & \cellcolor{tabblue!50}+0.35 & \cellcolor{tabblue!47}+0.33 & \cellcolor{tabamber!35}+0.21 & \cellcolor{tabamber!40}+0.24 \\
  & MaskGIT & 0.59 & 0.59 & 0.62 & 0.62 & \cellcolor{tabblue!4}+0.03 & \cellcolor{tabblue!4}+0.03 & \cellcolor{tabamber!0}+0.00 & \cellcolor{tabamber!2}+0.01 \\
\cmidrule(lr){2-10}
 \multirow{4}{*}{\pong} & Dreamer & 0.88 & 0.88 & 0.95 & 0.95 & \cellcolor{tabblue!10}+0.07 & \cellcolor{tabblue!10}+0.07 & \cellcolor{tabpurple!30}-0.00 & \cellcolor{tabpurple!30}-0.03 \\
  & AR-Trans. & 0.76 & 0.85 & 0.88 & 0.87 & \cellcolor{tabblue!16}+0.11 & \cellcolor{tabblue!2}+0.01 & \cellcolor{tabamber!42}+0.25 & \cellcolor{tabpurple!30}-0.00 \\
  & D3PM & 0.77 & 0.76 & 0.80 & 0.79 & \cellcolor{tabblue!4}+0.03 & \cellcolor{tabblue!5}+0.04 & \cellcolor{tabamber!7}+0.04 & \cellcolor{tabamber!7}+0.04 \\
  & MaskGIT & 0.68 & 0.76 & 0.98 & 0.95 & \cellcolor{tabblue!42}+0.29 & \cellcolor{tabblue!28}+0.20 & \cellcolor{tabamber!41}+0.25 & \cellcolor{tabamber!20}+0.12 \\
\cmidrule(lr){2-10}
 \multirow{4}{*}{\racer} & Dreamer & 0.68 & 0.69 & 0.71 & 0.75 & \cellcolor{tabblue!5}+0.04 & \cellcolor{tabblue!9}+0.06 & \cellcolor{tabamber!1}+0.00 & \cellcolor{tabpurple!30}-0.02 \\
  & AR-Trans. & 0.50 & 0.53 & 0.74 & 0.73 & \cellcolor{tabblue!35}+0.24 & \cellcolor{tabblue!29}+0.20 & \cellcolor{tabamber!47}+0.28 & \cellcolor{tabamber!28}+0.17 \\
  & D3PM & 0.62 & 0.62 & 0.64 & 0.64 & \cellcolor{tabblue!3}+0.02 & \cellcolor{tabblue!2}+0.01 & \cellcolor{tabamber!6}+0.03 & \cellcolor{tabpurple!30}-0.03 \\
  & MaskGIT & 0.62 & 0.61 & 0.64 & 0.63 & \cellcolor{tabblue!3}+0.02 & \cellcolor{tabblue!3}+0.02 & \cellcolor{tabamber!3}+0.02 & \cellcolor{tabamber!4}+0.02 \\
\cmidrule(lr){2-10}
 \multirow{4}{*}{\snake} & Dreamer & 0.10 & 0.10 & 0.61 & 0.64 & \cellcolor{tabblue!73}+0.51 & \cellcolor{tabblue!77}+0.54 & \cellcolor{tabamber!59}+0.36 & \cellcolor{tabamber!74}+0.44 \\
  & AR-Trans. & 0.10 & 0.10 & 0.59 & 0.83 & \cellcolor{tabblue!70}+0.49 & \cellcolor{tabblue!100}+0.74 & \cellcolor{tabamber!62}+0.37 & \cellcolor{tabamber!72}+0.43 \\
  & D3PM & 0.47 & 0.50 & 0.53 & 0.54 & \cellcolor{tabblue!8}+0.05 & \cellcolor{tabblue!5}+0.04 & \cellcolor{tabpurple!30}-0.01 & \cellcolor{tabpurple!30}-0.00 \\
  & MaskGIT & 0.50 & 0.43 & 0.53 & 0.48 & \cellcolor{tabblue!5}+0.03 & \cellcolor{tabblue!7}+0.05 & \cellcolor{tabpurple!30}-0.15 & \cellcolor{tabamber!0}+0.00 \\
\bottomrule
\end{tabular}
}
\end{table}

\subsection{Coding agents reliably improve base world models}
\label{sec:results-reliable-improvement}

Table~\ref{tab:cell-matrix} summarizes the experimental results, showing that both agents improve on the base in the large majority of sessions. Across the $64$ sessions, the agent-produced model exceeds the base on the held-out test split in $63$, with a mean test-score lift of $+0.196$ on a $[0,1]$ scale (median $+0.115$). The gains are not uniform: of the $63$ positive sessions, $33$ are substantial ($\Delta\ge+0.10$), $10$ moderate ($+0.05\le\Delta<+0.10$), and $20$ marginal ($0<\Delta<+0.05$), with $11$ of the marginal sessions marginal on both the test and scenario axes. Roughly two-thirds of the improvements are thus meaningful and about a third are marginal but real. The single exception is a \opusfour session on \breakout/D3PM that regresses by $0.001$ test score; \gptfivefour improves the base on every task. The magnitude of the lift tracks base strength: gains concentrate on tasks whose bases are weakest (\Cref{fig:game-hardness}), with \asteroids bases averaging $0.213$ and reaching $0.618$ ($+0.405$) while \pong bases begin at $0.792$ and reach $0.895$ ($+0.103$). The table carries per-agent Base columns because base re-runs across sessions produce different step counts and long-horizon rollout behavior despite identical code; platform non-determinism on the shared $8$-GPU host is the cause, and \Cref{sec:supp-starter-repro} documents it.

The same pattern holds on the scenario suite, a held-out evaluation over per-game rule probes (\Cref{sec:appendix-scenarios}). The scenario score uses the test score's per-horizon composite, but replaces $h_{20}$ with $h_{\mathrm{end}}$ (full rollout length), tightening the test of long-horizon rule consistency. With this weighting, the agent-best scenario score beats the base on $56$ of $64$ sessions, with a mean lift of $+0.170$ (median $+0.149$). Scenario and test lifts agree in direction on $55$ of $64$ sessions (task-level correlation $r = 0.89$; Table~\ref{tab:cell-matrix}, scenario columns); breakdowns and horizon decompositions are in \Cref{sec:supp-scenarios}.

\subsection{The two agents are not distinguishable at this sample size}
\label{sec:results-inter-agent}

\begin{figure}[t]
\centering
\begin{subfigure}[b]{0.32\linewidth}
  \includegraphics[width=\linewidth]{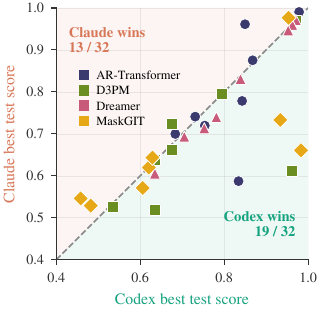}
  \caption{}
  \label{fig:agent-headtohead-scatter}
\end{subfigure}\hfill
\begin{subfigure}[b]{0.33\linewidth}
  \includegraphics[width=\linewidth]{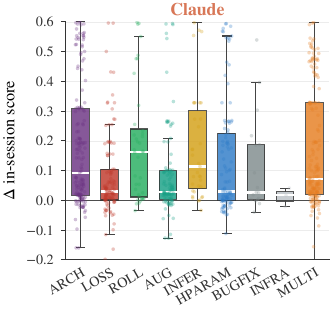}
  \caption{}
  \label{fig:agent-headtohead-uplift-opus}
\end{subfigure}\hfill
\begin{subfigure}[b]{0.33\linewidth}
  \includegraphics[width=\linewidth]{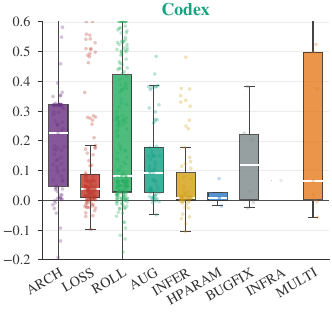}
  \caption{}
  \label{fig:agent-headtohead-uplift-codex}
\end{subfigure}
\caption{(\subref{fig:agent-headtohead-scatter}) Per-task best test score, marker shape and color denote the base architecture. Axes are zoomed to $[0.4, 1.0]$ because every observed $(x,y)$ pair falls in that range. Points above the $y{=}x$ diagonal are \opusshort wins. Split is $19$--$13$ in favor of \gptshort. (\subref{fig:agent-headtohead-uplift-opus}, \subref{fig:agent-headtohead-uplift-codex}) Per-experiment $\Delta$ in-session score across the full pool of $1{,}335$ experiments (validation composite observed by the agent during its session), grouped by \geminipronew{} change-type label. Boxes show the interquartile range, whiskers the $5$th and $95$th percentiles, and jittered points individual experiments. }
\label{fig:agent-comparison}
\end{figure}

Of the 32 tasks, \gptfivefour achieves the higher held-out test score on $19$ and \opusfour on $13$ (\Cref{fig:agent-headtohead-scatter}). Averaged over the same $32$ tasks, \opusfour reaches a mean best test score of $0.736$ and \gptfivefour $0.772$, a gap of $+0.036$, consistent in direction with the $19$--$13$ head-to-head split.

A paired Wilcoxon signed-rank test yields $W = 187$, $p = 0.15$ (median margin $+0.005$). At this sample size the point estimate favors \gptfivefour, but we do not detect a statistically significant difference between the two agents; demonstrating equivalence would require a pre-specified margin and an equivalence test (e.g., TOST), which we did not perform.

\paragraph{Caveat on best-of-$k$.} \opusfour runs a mean of $23.5$ experiments per session (median $25$) and \gptfivefour $18.2$ (median $19.5$). Because each task uses a best-of-$k$ test score and $k$ differs by agent, the ``more draws yield a higher maximum'' null is also consistent with the data: \gptfivefour may be stronger per experiment but penalized by $23\%$ fewer attempts, yielding the same $p = 0.15$ aggregate head-to-head result. The per-session distribution of $k$ is in \Cref{sec:supp-budget}; the time-to-best analysis is in \Cref{sec:supp-time-to-best}.

On our benchmark, the two agents reach comparable test scores, but \gptfivefour uses fewer tokens. Across sessions, \opusfour uses a median of $37.2$M total tokens ($36.9$M prompt, $354$k output), versus $25.9$M for \gptfivefour ($25.7$M prompt, $203$k output), giving \opusfour a $1.44\times$ higher token budget. Meanwhile, \gptfivefour achieves a slightly larger cumulative $\Delta$ test score over the $32$ shared tasks ($+6.71$ vs.\ $+5.85$). Together, these measurements suggest that \gptfivefour is about $1.65\times$ more token-efficient per unit score gain, a system-level comparison since the two agents differ in harness as well as model (\Cref{sec:limitations}). Full breakdowns and token statistics are in \Cref{sec:supp-token-efficiency}.

\subsection{Winning modifications are mainly non-trivial research edits, not hyperparameter sweeps}
\label{sec:results-nontrivial}

To characterize the edits produced during each session, we classify every experiment's change against its base using \geminipronew{} as a zero-shot judge (full methodology in \Cref{sec:supp-classifier}). The judge receives a unified diff of \texttt{config.json} and \texttt{train.py} together with the agent's optional \texttt{EXPERIMENT.md} writeup and returns one of nine labels: \textsc{architecture}, \textsc{loss}, \textsc{rollout}, \textsc{inference}, \textsc{data\_aug}, \textsc{hyperparam}, \textsc{bugfix}, \textsc{infra}, or \textsc{multiple} (assigned when three or more distinct categories are active with no stated primary intent). For the aggregate analysis we group \textsc{hyperparam} and \textsc{infra}---pure scalar-knob or plumbing changes---as \emph{trivial} and treat every other label as \emph{non-trivial}. Table~\ref{tab:change-type} reports the full nine-label breakdown. The trivial/non-trivial split is near-mechanical and reproduces at $94.6\%$ when the agent's stated intent is stripped and only the code diff is shown (\Cref{sec:supp-classifier-ablation}); the fine nine-label categories, by contrast, depend on the stated intent, so we present the per-category breakdown as descriptive rather than as an objective code classification.

The experiments that win each session are predominantly non-trivial; in $58$ of $64$ sessions ($91\%$), the session-best experiment is non-trivial, while the remaining six sessions are won by hyperparameter-only modifications (\Cref{tab:change-type}).
Across the full pool of $1{,}335$ classified experiments, the non-trivial subset ($n = 1{,}220$) achieves a mean $\Delta$ in-session score of $+0.146$ and exceeds the base on $85\%$ of experiments, compared with $+0.126$ mean $\Delta$ and a $73\%$ win rate for the $n = 115$ trivial subset. We characterize novelty qualitatively rather than by the automatic novelty flag, which is intent-sensitive like the fine labels: an audit of the flagged experiments finds a small number of genuinely non-obvious, code-verified mechanisms---for example, hand-crafted kinematic motion priors and toroidal wrapped-delta state representations---against a larger set of recurring standard archetypes.

The per-label uplift distributions in \Cref{fig:agent-headtohead-uplift-opus,fig:agent-headtohead-uplift-codex} further indicate that the two agents are similarly effective per category: both extract positive median in-session lift from every non-trivial label (\textsc{loss}, \textsc{architecture}, \textsc{rollout}, \textsc{inference}, \textsc{data\_aug}), and the apparent per-agent differences---\gptfivefour's wider right tail on \textsc{rollout}, \opusfour's wider right tail on \textsc{multiple}---reflect how each agent allocates its session budget across categories rather than a per-category skill gap.

The winning edits are structural, not numerical: seven of the eight largest per-game lifts add objectives, representation changes, rollout-time procedures, or architectural edits; only \pong/MaskGIT uses a learning-rate and step-budget schedule (per-game case studies in \Cref{tab:case-studies}). Across sessions these edits recur as a few mechanism families, all attacking compounding autoregressive error at long horizons (\Cref{sec:supp-case-studies}).

Every winning edit modifies \texttt{train.py} and so falls outside the base's fixed hyperparameter space, which is why knob-only search does not reach it; \Cref{sec:results-controls} makes this precise against compute-matched search baselines.
The resulting training trajectories appear in the per-experiment validation-loss curves in \Cref{sec:supp-valloss}, and a descriptive per-category selectivity decomposition in \Cref{sec:supp-label-selection}.

\subsection{Gains concentrate at long-horizon rollout, not one-step prediction}
\label{sec:results-horizon}

The test score combines three horizon composites: $0.1 \cdot c_1 + 0.2 \cdot c_{10} + 0.7 \cdot c_{20}$. Decomposing mean test-score lift across them shows whether agents improve one-step fit or long-horizon rollout behavior. Table~\ref{tab:horizon-breakdown} reports paired base-vs-agent means by horizon over the $64$ sessions.

\begin{table}[H]
\centering
\small
\caption{Per-horizon decomposition of the mean test-score lift, over the $64$ sessions. Base and agent-best values are per-horizon composites $c_h$; $\Delta$ is the paired per-session difference. The rightmost column reports how many of the $64$ sessions improve at that horizon.}
\label{tab:horizon-breakdown}
\begin{tabular}{l r r r r}
\toprule
Horizon & Base & Agent-best & $\Delta$ & $\#$ sessions with $\Delta > 0$ \\
\midrule
$h = 1$ & 0.808 & 0.864 & $+0.056$ & 47/64 \\
$h = 10$ & 0.583 & 0.788 & $+0.205$ & 62/64 \\
$h = 20$ & 0.522 & 0.737 & $+0.215$ & 63/64 \\
\bottomrule
\end{tabular}

\end{table}

The effect is horizon-asymmetric. At $h_1$, the base is already strong (mean $0.808$), with only a $+0.056$ average lift, positive on $47$ of $64$ sessions. At $h_{10}$ and $h_{20}$, the base drops to $0.583$ and $0.522$, while the agent recovers performance, with mean lifts of $+0.205$ and $+0.215$, positive on $62$ and $63$ of $64$ sessions---nearly every session improves long-horizon rollout. Agents are therefore not raising test scores by tuning one-step fit; they produce models that roll out state more accurately under their own predictions. The same asymmetry holds on the scenario split, including at $h_{\mathrm{end}}$~(\Cref{tab:scenario-horizon-breakdown}).

\subsection{The gains survive search baselines, re-seeding, and a reward-hacking audit}
\label{sec:results-controls}

Three checks test whether these gains reflect genuine research rather than easy tuning, seed luck, or scorer gaming. Against the compute-matched search baselines, the best agent produces the better model on the large majority of cells---beating random search on $27$ of $32$ and TPE on $25$ of $32$ on the held-out scenario suite ($28$--$0$--$4$ and $26$--$2$--$4$ on the test split), and closing about twice the headroom random search closes (median $46\%$ vs.\ $21\%$)---so knob tuning alone does not reach the agents' edits, which modify \texttt{train.py} and lie outside the searched space (\Cref{sec:supp-search-baseline}). The gains also survive re-seeding: $94$ of the $96$ repeated sessions improve on their base, with a median improvement ($+0.14$) roughly $2.9\times$ the median run-to-run seed spread ($0.049$); the magnitude is game-dependent and should be read per game (\Cref{sec:supp-seed-robustness}). Finally, an audit of all $64$ sessions against the safeguards of \Cref{sec:harness} finds no scorer tampering---self-reported and independent scores diverge by a median of only $0.008$---and a few disclosed boundary cases that all remain positive on the held-out test split (\Cref{sec:appendix-audit}).

\section{Limitations}
\label{sec:limitations}
Our results measure complete coding-agent systems under a fixed wall-clock budget. The \gptfivefour--\opusfour comparison therefore includes both model and harness: context management, restart behavior, tool use, and experiment scheduling. Separating model quality from harness effects would require swapping harnesses or evaluating both models under a shared orchestration layer. We therefore read the head-to-head results as system-level comparisons and avoid model-level claims; in particular, the token-efficiency observation (\Cref{sec:results-inter-agent}) reflects harness differences (a restart-loop versus a single continuous session) as much as any per-token model difference, and should be read as caveated and system-level.

The primary score emphasizes action-conditioned rollout fidelity, especially at $h_{10}$ and $h_{20}$. Thus, aggregate-score gains should not be read as uniform gains at every horizon. One-step prediction is already strong for many bases, while most gains occur in longer open-loop rollouts. The scenario suite adds held-out curated states that isolate game mechanics, evaluated with the same per-horizon composite but with the long-horizon term at $h_{\mathrm{end}}$; it still scores state fidelity rather than hand-written symbolic predicates for each rule event.

Our benchmark uses explicit structured state from game engines.  This removes perception, gives exact targets, and enables fast closed-loop experiments, but limits the benchmark to transition modeling over known entities. Architectures or recipes found here may need adaptation for pixel-based settings. Likewise, the current tensorization covers fields needed by our eight games; games depending on visual attributes or omitted physics variables would require extending the representation. We also emphasize that this is a deliberately clean substrate: the state is exact, with no sensor noise, missing values, or measurement error, whereas real scientific data typically contains all three. The tensor format naturally supports noise-injection and masking variants, which we see as the right extension once the clean-substrate measurement is established.

Finally, the reported per-cell magnitudes carry seed sensitivity. Most sessions were run once; the seed-repetition study (\Cref{sec:results-controls}) finds the improvements robust to seed on the large majority of cells, but the magnitude varies on some (for example, Snake/\gptshort/Dreamer reaches best scores of $0.64$/$0.92$/$0.65$ across three seeds), and on already-strong bases such as several Racer cells the gains are small and comparable to the seed spread. The direction of improvement is robust where there is headroom; the size should be read per game, not as a single blanket effect.

Our evaluation targets the world-modeling aspects with exact ground truth: action-conditioned state prediction, long-horizon rollout consistency, and scenario-level state evolution. This separates world-model quality from choices about planners, policies, rewards, or downstream controllers.

\section{Conclusion}
\label{sec:conclusion}

{We presented \datasetname{}, a closed-loop benchmark that evaluates AI coding agents as automated world-model researchers across eight games, four base architectures, and a fixed compute budget. Both \gptfivefour and \opusfour reliably improve their provided base, and the improvement is concentrated in long-horizon rollout rather than one-step fit. Unlike engineering-to-spec benchmarks that specify the direction of improvement in advance, \datasetname{} measures whether agents can steer research-style exploration under a compute cap; we release the benchmark, base code, and per-experiment artifacts to support further study.}

\bibliographystyle{plainnat}
\bibliography{references}

\clearpage
\appendix
\appendixtitleblock

\appendixphasetrue

\setcounter{figure}{0}
\setcounter{table}{0}
\renewcommand{\thefigure}{\thesection.\arabic{figure}}
\renewcommand{\thetable}{\thesection.\arabic{table}}

\section{Data Schema and Tensor Representation Details}
\label{sec:appendix-schema}

This appendix provides concrete examples and full specifications for the structured-state frame envelope (\Cref{sec:ecs}) and the data tensor representation (\Cref{sec:tensor}).

\subsection{Frame Envelope Example}
\label{sec:appendix-envelope}

\begin{figure}[H]
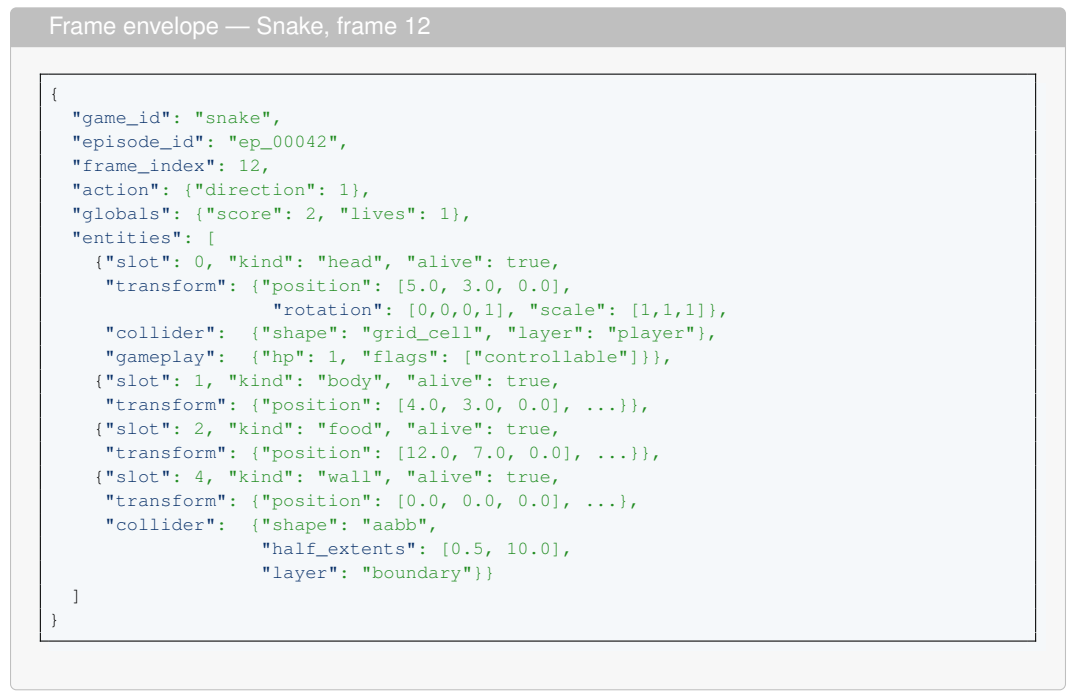

\centering
\begin{tcolorbox}[
  colback=black!3, colframe=black!25, fontupper=\ttfamily\fontsize{7}{9}\selectfont,
  title={\sffamily\small Frame envelope --- Snake, frame 12},
  width=\linewidth, boxrule=0.4pt, arc=2pt
]
\begin{lstlisting}[language=json, basicstyle=\ttfamily\fontsize{7}{9}\selectfont, numbers=none]
{
  "game_id": "snake",
  "episode_id": "ep_00042",
  "frame_index": 12,
  "action": {"direction": 1},
  "globals": {"score": 2, "lives": 1},
  "entities": [
    {"slot": 0, "kind": "head", "alive": true,
     "transform": {"position": [5.0, 3.0, 0.0],
                    "rotation": [0,0,0,1], "scale": [1,1,1]},
     "collider":  {"shape": "grid_cell", "layer": "player"},
     "gameplay":  {"hp": 1, "flags": ["controllable"]}},
    {"slot": 1, "kind": "body", "alive": true,
     "transform": {"position": [4.0, 3.0, 0.0], ...}},
    {"slot": 2, "kind": "food", "alive": true,
     "transform": {"position": [12.0, 7.0, 0.0], ...}},
    {"slot": 4, "kind": "wall", "alive": true,
     "transform": {"position": [0.0, 0.0, 0.0], ...},
     "collider":  {"shape": "aabb",
                   "half_extents": [0.5, 10.0],
                   "layer": "boundary"}}
  ]
}
\end{lstlisting}
\end{tcolorbox}
\caption{A complete structured-state frame envelope for \snake.
Mutable entities (head, body, food) are prediction targets; immutable entities (wall) serve as conditioning context.
Components attach optionally---the wall carries no Physics or Gameplay component.}
\label{fig:envelope-example}
\end{figure}

\Cref{fig:envelope-example} shows a complete frame envelope for \snake at frame~12.
The top-level fields record the game identifier, frame index, action, and global state.
The \texttt{entities} array contains one entry per occupied slot; here we show the head entity (slot~0, mutable) and a wall segment (slot~4, immutable).
Each entity carries only the components relevant to its role---the wall has no Physics component because it never moves.

\subsection{Data Tensor Layout}
\label{sec:appendix-formatd}

The data tensor representation converts each episode into three tensors.
\Cref{tab:formatd-layout} summarizes the layout; the subsections below walk through a concrete example.

\begin{table}[H]
\centering
\caption{Data tensor layout. $N$ = entity slot budget; $T$ = episode length.}
\label{tab:formatd-layout}
\small
\begin{tabular}{llll}
\toprule
\textbf{Tensor} & \textbf{Shape} & \textbf{Computed} & \textbf{Contents} \\
\midrule
Registry $\mathbf{R}$  & $[N, 34]$ & Once/episode & Collider shape (4), radius (1), half-ext.\ (2), \\
                        &                       &              & layer (18), trigger (1), mutable (1), scale (2), \\
                        &                       &              & + physics material (5) \\
\addlinespace
State $\mathbf{S}_t$   & $[T, N, 23]$          & Per frame   & pos\_norm (2), alive (1), vel\_norm (2), \\
                       &                       &              & gameplay (14, masked), pos\_history (4) \\
\addlinespace
Action $\mathbf{a}_t$  & $[T, 7]$   & Per frame   & player input (unified, per-game masked) \\
Game state $\mathbf{g}_t$  & $[T, 17]$  & Per frame   & scores, lives, counters (unified, per-game masked) \\
Terminal $t_t$ & $[T]$    & Per frame   & episode termination flag \\
\bottomrule
\end{tabular}
\end{table}

\subsec{Worked example: \pong (200 frames, 5 entities).}
\pong has $N{=}5$ entity slots: \texttt{paddle\_l} (mutable), \texttt{paddle\_r} (mutable), \texttt{ball} (mutable), \texttt{wall\_top} (immutable), \texttt{wall\_bottom} (immutable).
All 5 slots are alive in every frame.

\begin{compactitem}
\item \textbf{Registry} $\mathbf{R} \in \mathbb{R}^{5 \times 34}$: Slot~0 (\texttt{paddle\_l}) encodes \texttt{aabb} shape $=[1,0,0,0]$, half-extents $=[0.013, 0.083]$, mutable $=1$.
Slot~2 (\texttt{ball}) encodes \texttt{circle} shape $=[0,1,0,0]$, radius $=0.021$, restitution $=1.0$.
Slots~3--4 (walls) have mutable $=0$, half-extents $=[0.5, 0.004]$.
\item \textbf{Dynamic state} $\mathbf{S} \in \mathbb{R}^{T \times 5 \times 23}$: $D_s{=}23$ (5~base fields + 14~gameplay + 4~position history deltas).  At frame~$t$, slot~2 (ball) reads $[0.50, 0.50, 1.0, {-}0.011, {-}0.014, 0, \ldots, 0]$---normalized position, alive $=1$, velocity, and 18~additional fields (zeroed for \pong, which has no gameplay state).
\item \textbf{Actions and globals:} Actions $\in \mathbb{R}^{T \times 7}$ (\pong uses 2 of 7 dims: \texttt{paddle\_l\_dy}, \texttt{paddle\_r\_dy}); globals $\in \mathbb{R}^{T \times 17}$ (score\_l, score\_r, and masked fields).
\end{compactitem}

\noindent\textbf{Size comparison.}
The data tensor for a 200-frame \pong episode uses $5 \times 34 + 200 \times 5 \times 23 + 200 \times (7 + 17) = \num{170} + \num{23000} + \num{4800} = \num{27970}$ floats---compact because only 5~entity slots are needed.

\subsection{Cross-Game Registry Structure}
\label{sec:appendix-crossgame}

All games share the same 34-dimensional registry schema, so structurally similar entities across games occupy the same feature space even though their specific values differ.
For example, \texttt{ball} in \pong and \breakout both encode the \texttt{circle} shape one-hot, a positive radius, and the mutable flag---but differ in exact radius (0.021 vs.\ 0.017) and scale due to different arena dimensions.
Similarly, paddles in both games share the \texttt{aabb} shape and mutability but differ in half-extents and collision-layer assignment.
This shared schema means a model that conditions on registry vectors implicitly receives comparable structural cues across games, even without an explicit game identifier.

\subsection{Per-Game Entity and Action Details}
\label{sec:appendix-pergame}

\Cref{tab:pergame-entities} provides the full entity roster for each game, including kind labels, mutability, and the action fields used.

\begin{table}[H]
\centering
\caption{Per-game entity kinds and action fields.
Bold entity kinds are immutable (conditioning only).
Action fields map into a unified 7-dimensional action vector via per-game masks.}
\label{tab:pergame-entities}
\small
\begin{tabular}{lp{5.5cm}l}
\toprule
\textbf{Game} & \textbf{Entity Kinds} & \textbf{Action Fields} \\
\midrule
Snake          & head, body, food, \textbf{wall}                               & direction (discrete, 4 classes) \\
Frogger        & frog, car, log, \textbf{goal}, \textbf{water}                 & action (discrete, 5 classes) \\
Pong           & paddle\_l, paddle\_r, ball, \textbf{wall\_top}, \textbf{wall\_bottom} & paddle\_l\_dy, paddle\_r\_dy (continuous) \\
Breakout       & paddle, ball, brick                                           & action (discrete, 3 classes) \\
Asteroids      & ship, bullet, asteroid\_lg/md/sm                              & action (discrete, 8 classes) \\
Platformer     & player, coin, \textbf{platform}, \textbf{spike}, \textbf{goal} & jump, move\_x (continuous) \\
Kong           & player, barrel, item, \textbf{ladder}, \textbf{platform}      & jump, move\_x, move\_y (continuous) \\
Racer          & car, obstacle, \textbf{lane\_marker}                          & move\_x (continuous) \\
\bottomrule
\end{tabular}
\end{table}

\Cref{tab:action-schema} lists the exact byte-level action schema: which slot of the $7$-dim vector each game writes to, the type of the value stored there, and the range (continuous) or class list (discrete). The slot indices and ranges below were enumerated from every episode of the test split; all other slots are zero for that game and masked out by the per-frame \texttt{action\_masks} field.

\begin{table}[H]
\centering
\caption{Per-game action-tensor schema, verified against the test split. Slot indices $0$--$6$ refer to the shared $7$-dim action vector $\mathbf{a}_t \in \mathbb{R}^{7}$; a game populates only the slots listed here.}
\label{tab:action-schema}
\small
\begin{tabular}{l c l l l}
\toprule
\textbf{Game} & \textbf{Slot} & \textbf{Field name} & \textbf{Type} & \textbf{Range / classes} \\
\midrule
Snake      & $1$ & \texttt{direction}    & discrete   & $\{0,1,2,3\}$ \\
Frogger    & $0$ & \texttt{action}       & discrete   & $\{0,1,2,3,4\}$ \\
\addlinespace
Pong       & $5$ & \texttt{paddle\_l\_dy} & continuous & $[-1,+1]$ \\
           & $6$ & \texttt{paddle\_r\_dy} & continuous & $[-1,+1]$ \\
\addlinespace
Breakout   & $0$ & \texttt{action}       & discrete   & $\{0,1,2\}$ \\
Asteroids  & $0$ & \texttt{action}       & discrete   & $\{0,1,2,3,4,5,6,7\}$ \\
\addlinespace
Platformer & $2$ & \texttt{jump}         & continuous & $[0,1]$ \\
           & $3$ & \texttt{move\_x}      & continuous & $[-1,+1]$ \\
\addlinespace
Kong       & $2$ & \texttt{jump}         & continuous & $[-1,+1]$ \\
           & $3$ & \texttt{move\_x}      & continuous & $[-1,+1]$ \\
           & $4$ & \texttt{move\_y}      & continuous & $[-1,+1]$ \\
\addlinespace
Racer      & $3$ & \texttt{move\_x}      & continuous & $[-1,+1]$ \\
\bottomrule
\end{tabular}
\end{table}

The \texttt{action} slot (index $0$) is a shared discrete field used by games that pack several controller inputs into one class index (Atari-style minimal action sets): Frogger's $5$-class field covers the four directions plus a no-op; Breakout's $3$-class field is (no-op, left, right); Asteroids' $8$-class field covers combinations of thrust, rotate and fire. Platformer's \texttt{jump} is an analog magnitude in $[0,1]$ rather than a binary flag; Kong's \texttt{jump} is signed in $[-1,+1]$. These are the values a world model receives and must condition on; the semantic labels (\texttt{thrust/rotate/shoot}, \texttt{steer/accelerate}, \texttt{climb}) used in some prior drafts are descriptions of the underlying game controls, not separate tensor fields.

\subsection{Reward Shaping for RL Data Collection}
\label{sec:appendix-reward-shaping}

Six games include RL-generated trajectories alongside heuristic and random data.
\snake and \frogger use DQN~\citep{mnih2015dqn}; \pong, \kong, \platformer, and \racer use PPO~\citep{schulman2017ppo}.
All RL agents are trained with potential-based reward shaping~\citep{ng1999shaping} to accelerate convergence, with shaping magnitudes kept 5--30$\times$ smaller than task rewards to avoid dominating the gradient.
The shaped rewards are applied only during RL training and do not appear in the collected dataset.

\paragraph{\snake.} Distance-to-food shaping: the agent receives a small positive reward proportional to the decrease in Manhattan distance between the head and the food ($+0.1 \times \Delta d / d_{\max}$), plus a constant survival penalty ($-0.005$ per step) to discourage stalling.

\paragraph{\pong.} Paddle--ball alignment: the active paddle (the one the ball is moving toward) receives a bonus proportional to its vertical alignment with the ball ($+0.02 \times \text{alignment}$).  The idle paddle receives a smaller center-positioning bonus ($+0.005$) to encourage returning to a neutral position.

\paragraph{\racer.} Survival bonus ($+0.02$ per step), road-centering reward ($+0.01 \times \text{centering}$), and enemy-clearance bonus ($+0.01$) when an oncoming car is nearby, encouraging the agent to maintain safe lateral distance.

\paragraph{\frogger, \kong, \platformer.} These games use the base game reward without additional shaping terms, as the native reward signal proved sufficient for PPO/DQN convergence.

All RL agents are trained for 10M environment steps with default PPO/DQN hyperparameters (learning rate $3 \times 10^{-4}$, 8 parallel environments for PPO).

\subsection{Dataset Statistics}
\label{sec:appendix-dataset}

\Cref{tab:dataset-stats} reports per-game maximum entity count, total frames, average episode length, and state dimensionality; every game contributes a uniform $19{,}000$ episodes.
The dataset totals 152{,}000 episodes and nearly 158 million frames across eight games.
Episode lengths vary by over an order of magnitude: short games (\asteroids, \frogger) average ${\sim}300$ frames, while long-rally games (\pong, \breakout) average over 1{,}700 frames.

\begin{table}[H]
\centering
\caption{Per-game dataset statistics. Each game contributes $19{,}000$ episodes divided into the four splits defined in \Cref{sec:exp-setup} (training, validation, test, scenario).}
\label{tab:dataset-stats}
\small
\setlength{\tabcolsep}{4pt}
\begin{tabular}{lrrrr}
\toprule
\textbf{Game} & \textbf{Max Ent.} & \textbf{Total Frames} & \textbf{Avg Len} & \textbf{$D_s$} \\
\midrule
Snake          &  48 &  15{,}910{,}306 &   837 & 23 \\
Frogger        &  28 &  6{,}247{,}436 &   329 & 23 \\
Pong           &   5 &  42{,}499{,}735 & 2{,}237 & 23 \\
Breakout       &  52 &  33{,}042{,}611 & 1{,}739 & 23 \\
Asteroids      &  20 &   5{,}635{,}042 &   297 & 23 \\
Platformer     &  24 &  10{,}572{,}906 &   556 & 23 \\
Kong           &  16 &  22{,}966{,}344 & 1{,}209 & 23 \\
Racer          &   6 &  21{,}119{,}547 & 1{,}112 & 23 \\
\midrule
\textbf{Total} &     & \textbf{157{,}993{,}927} & & \\
\bottomrule
\end{tabular}
\end{table}

\noindent All games share the same state dimensionality ($D_s{=}23$: 2 position + 1 alive + 2 velocity + 14 gameplay fields + 4 position history deltas) and registry dimensionality (34).
The number of active gameplay fields varies per game (masked to zero for unused fields).
Entity counts range from 5 (\pong: fixed roster of paddles, ball, walls) to 52 (\breakout: paddle, ball, and up to fifty bricks), directly affecting transformer-based model memory requirements.

\clearpage
\section{Base Model Architectures}
\label{sec:appendix-models}

This section details the four base architectures introduced in \Cref{sec:exp-setup}. All four base models share the same input/output contract: given a window of $W{=}8$ historical frames (registry, states, actions, globals), predict next-frame entity positions, alive status, terminal flag, and gameplay fields for mutable entities.
\Cref{tab:model-hyperparams} summarizes the key hyperparameters.

\begin{table}[H]
\centering
\caption{Base model hyperparameters (shared across all games).}
\label{tab:model-hyperparams}
\small
\begin{tabular}{lcccc}
\toprule
& \textbf{Dreamer} & \textbf{AR-Trans.} & \textbf{D3PM} & \textbf{MaskGIT} \\
\midrule
Hidden dim        & 256    & 256    & 256    & 256 \\
Attention heads   & ---    & 8      & 8      & 8 \\
Encoder layers    & ---    & 3      & 3      & 3 \\
Decoder layers    & ---    & 1 (cross-attn) & 1 (cross-attn) & 1 (cross-attn) \\
Dropout           & ---    & 0.1    & 0.1    & 0.1 \\
Max frames        & ---    & 9 (W+1) & 9 (W+1) & 9 (W+1) \\
Latent dim        & $32 \times 32$ categorical & ---    & ---    & --- \\
Diffusion steps   & ---    & ---    & 100    & --- \\
Mask iterations   & ---    & ---    & ---    & 8 \\
Free nats (KL)    & 1.0    & ---    & ---    & --- \\
Vel.\ consistency & 0.1    & 0.1    & 0.1    & 0.1 \\
Learning rate     & 3e-4   & 3e-4   & 3e-4   & 3e-4 \\
\bottomrule
\end{tabular}
\end{table}

\paragraph{Dreamer (RSSM).}
The recurrent model follows DreamerV3~\citep{hafner2023mastering} adapted for structured entity state.
An entity encoder projects each entity's concatenated registry and state vectors through a two-layer MLP, then mean-pools across entities to produce a global entity representation.
A separate context encoder processes the concatenated action and global vectors.
Both are fed into a GRU cell that maintains a hidden state $h \in \mathbb{R}^{256}$ across timesteps.
From $h$, a prior network and a posterior network each produce logits for 32 categorical variables with 32 classes, yielding a discrete latent $z \in \{0,1\}^{32 \times 32}$ via straight-through sampling.
Decoding is per-entity: the concatenation of $h$, $z$, and the entity's registry vector is projected to predict position deltas (via symlog transform), alive logits, terminal logit (from pooled $hz$), and gameplay fields.
Training uses symlog MSE for positions, binary cross-entropy for alive/terminal, and free-bits KL regularization ($\text{free\_nats}{=}1.0$) between prior and posterior.
At inference, only the prior is used (no access to future states).

\paragraph{AR-Transformer.}
A continuous autoregressive Transformer inspired by IRIS~\citep{micheli2023transformers}, but operating on continuous entity state rather than discrete image tokens.
Each entity in each frame is projected via a two-layer MLP from the concatenation of its registry and state vectors, plus temporal position embeddings and action/global embeddings.
The resulting entity tokens for $W{+}1$ frames (W context + 1 target) are encoded by a block-causal TemporalContextEncoder: within each frame, entities attend bidirectionally; across frames, attention is causal (each frame can only attend to itself and earlier frames).
A single-layer cross-attention decoder then produces per-entity predictions by attending target entity tokens to the encoded context.
Output heads predict position deltas, alive logits, terminal logit (from masked mean-pool), and gameplay fields.
Training uses symlog MSE for positions with the same velocity consistency regularization as Dreamer.

\paragraph{D3PM.}
A discrete denoising diffusion model following Austin et al.~\citep{austin2021d3pm}.
Entity states are first quantized via an EntityTokenizer that maps continuous positions to discrete bins ($K_x, K_y$ per axis, chosen per game to match the simulator's native position resolution rather than a single default --- ranging from $20\times 20$ for Snake's grid world to $256 \times 224$ for Frogger and $200 \times 200$ for Asteroids) and alive/gameplay fields to categorical tokens.
The same TemporalContextEncoder as AR-Transformer encodes $W{+}1$ context frames.
For each target frame, a diffusion timestep $t \sim \text{Uniform}(1, T{=}100)$ is sampled, target tokens are corrupted with uniform noise proportional to $\bar{\alpha}_t$, and a cross-attention decoder denoises them conditioned on the temporal context plus a time embedding.
Per-field classification heads predict the clean token.
At inference, reverse diffusion runs for 10 steps per frame, autoregressively appending each predicted frame to the context with a sliding window cap.

\paragraph{MaskGIT.}
A masked generative Transformer following Chang et al.~\citep{chang2022maskgit}.
Architecture matches D3PM (same tokenizer, TemporalContextEncoder, cross-attention decoder) but replaces the diffusion noise schedule with a cosine mask schedule.
During training, a random fraction of target entities (sampled from a cosine distribution, minimum 10\%) have all their token fields replaced with a learnable MASK token.
The decoder predicts the masked tokens conditioned on the unmasked tokens and the temporal context, with a mask-ratio embedding providing schedule information.
At inference, iterative parallel decoding runs for 8 iterations: at each iteration, the most confident predictions are unmasked and the remaining tokens are re-predicted, progressively filling in the full prediction.

\paragraph{Shared components.}
The three Transformer-based models share a common TemporalContextEncoder from a shared \texttt{temporal.py} module: a stack of 3 Transformer layers with 8-head attention and block-causal masking.
All four models share: (1) the velocity consistency loss that penalizes disagreement between predicted position deltas and the velocity implicit in consecutive positions ($\lambda{=}0.1$), (2) the same training infrastructure (AdamW, linear warmup over 500 steps, gradient clipping at 1.0, mixed-precision training), and (3) the same data pipeline producing windowed tensors at $W{=}8$ with stride 4.

\clearpage
\section{Agent Harness Details}
\label{sec:appendix-harness-details}

This section elaborates the closed-loop harness of \Cref{sec:harness}: the verbatim task instructions the agent receives, and how sessions persist across the two agents.

\subsection{Agent Task Instructions}
\label{sec:appendix-instruction}

Each task directory contains an \texttt{instruction.md} file that the agent reads at the start of its session.
Every session is launched from the same instruction template; the placeholders \texttt{\{Game\}}/\texttt{\{game\}}, \texttt{\{Model\}}/\texttt{\{model\}}, the \texttt{\{max\_entities\}} count, and the per-game state-schema fields (\texttt{\{active\_fields\}}, \texttt{\{core\_field\_count\}}, \texttt{\{gameplay\_field\_count\}}) are substituted at setup time from each base's \texttt{config.json} and the game's \texttt{meta.json}. The verbatim template below is the source of truth; the per-task values are mechanically derived. The template predates our terminology and refers to the agent's starting point as the ``starting-point world model'' or ``baseline''; this is the object the paper calls the \emph{base world model}.

The template encodes the research goal, the file-permission contract, the mechanical workflow (run the base first, then iterate), the scoring formula the verifier uses, and the experiment-tracking conventions. The \texttt{run.py} and \texttt{score.py} scripts referenced by the template are shipped with the task directory and are read-only during the session, so the agent can only influence the final score by editing \texttt{train.py} and \texttt{config.json}.

\medskip
\noindent\textbf{\texttt{instruction.md} (verbatim template).}
\vspace{2pt}

\begin{lstlisting}[%
  basicstyle=\ttfamily\fontsize{7}{8.4}\selectfont,
  commentstyle={},
  keywordstyle={},
  stringstyle={},
  identifierstyle={},
  emph={}, emphstyle={},
  showstringspaces=false,
  numbers=none,
  breaklines=true,
  breakatwhitespace=true,
  breakindent=0pt,
  postbreak=\mbox{\textcolor{gray}{$\hookrightarrow$}\space},
  frame=single,
  framerule=0.4pt,
  rulecolor=\color{black!40},
  backgroundcolor=\color{backcolor},
  xleftmargin=4pt,
  xrightmargin=4pt,
  linewidth=\linewidth,
  aboveskip=2pt,
  belowskip=2pt,
  keepspaces=true,
  columns=fullflexible,
  upquote=true,
]
# {Game} -- {Model} World Model

## Mission

Treat the {model} as a **starting-point world model** for {Game}.

The model receives a short history of past frames and must predict the **next
frame**. By repeating this process autoregressively, it should serve as an
**interactive forward model** of the game that can be stepped forward and
compared against the true simulator.

Your job is to improve the baseline so that predicted rollouts remain faithful to
real gameplay, especially over longer horizons where small errors compound.

## Success Criterion

The **only objective that matters** is the **final score** produced by the evaluator from a completed run.

Per-horizon composite:

`0.9 * (1 - position_l1) + 0.1 * alive_f1`

Final score:

`0.1 * h1 + 0.2 * h10 + 0.7 * h20`

This means:
- **90% of the final score comes from h10 and h20**.
- **Position accuracy dominates** each per-horizon composite.
- Single-step quality matters only insofar as it improves long-horizon rollout fidelity.

### Critical interpretation

- `val_loss`, training loss, wall-clock time, and throughput are **diagnostics only**.
- They are **not optimization targets**.
- A lower `val_loss` does **not** count as an improvement unless the **final score** improves.
- A shorter training run does **not** count as an improvement unless the **final score** improves in a controlled comparison.

When making decisions, optimize for **h10/h20 rollout stability**, not for one-step validation behavior.

## Data

- Game: {game} ({max_entities} max entities per frame)
- Active gameplay fields: {active_fields}
- Shared cache: `/data/{game}/_cache/`
- State: {core_field_count} core fields (pos_x, pos_y, alive, vel_x, vel_y) + {gameplay_field_count} gameplay fields

## Constraints

- Per-experiment training budget: **600s**
- Window size: **8** frames
- Evaluation horizons: **h=1, h=10, h=20**

## Ground Rules

### Files you may modify
You may ONLY modify:
- `train.py` -- model architecture, optimizer, hyperparameters, training loop
- `config.json`
- `configs/*` -- create new configs here
- `experiments/*` -- experiment outputs, logs, notes

Everything else is READ-ONLY. Do not modify, overwrite, or shadow:
- `score.py`
- `run.py`
- `evaluator.py`
- anything under `/opt/worldbench/lib/`
- `instruction.md`
- `/data/`

Do not create new Python files that shadow existing modules.
The evaluation harness is the ground-truth metric.

## First Action: Establish the Baseline

**Check `summary.tsv` first.** If it already has data rows, skip the baseline
and continue iterating from the latest result -- you are resuming a prior session.

If `summary.tsv` does not exist or has no data rows, your first run must be the
unmodified baseline:

```bash
python run.py --config config_template.json
```

Do not modify any code before this first run.
All future experiments must be judged against this baseline.

## Running Experiments

```bash
python run.py --config config_template.json
python run.py --config configs/my_experiment.json
```

Each run trains, evaluates, saves results under:

`experiments/{game}_{model}/<run_id>/`

After each run, write `EXPERIMENT.md` in that run directory describing:
- hypothesis
- exact change made
- why the change should help h10/h20 rollout fidelity
- final score and key horizon results
- what you learned

## Training Budget

Per-experiment training budget:
- **600 seconds maximum** from `config.json max_train_seconds`

- For any **serious, non-failing experiment**, use the **full allotted training budget**.
- Do **not** introduce early stopping as a default strategy.
- Do **not** shorten training just to run more experiments.
- Do **not** prefer earlier checkpoints because they have lower `val_loss`.

Early stopping is only allowed if the explicit hypothesis is that less training
improves the **final score**, tested as a controlled experiment against a full-budget run.

## Research Priorities

Prioritize changes that are plausibly causal for **long-horizon rollout fidelity**.

Good directions include:
- training objectives that better match open-loop rollout behavior
- methods that reduce compounding error across steps
- architecture changes that improve temporal consistency and entity dynamics
- losses or curricula that emphasize h10/h20 behavior
- better handling of velocity, persistence, alive/dead transitions, and structured state evolution
- training strategies that improve robustness under autoregressive rollout

When choosing between ideas, prefer the one with the stronger causal connection to:
- lower position drift over long horizons
- more stable open-loop rollouts
- better entity persistence and transition modeling

## Experiment Validity

A run only counts as evidence if:
1. It uses the standard evaluation harness.
2. It completes evaluation and produces a **final score**.
3. It is compared fairly against prior runs.
4. Its claimed improvement is based on **final score**, not on proxy metrics.

Never prefer an experiment because it has a better `val_loss` if its **final score** is worse or unproven.

Keep comparisons fair -- unless explicitly testing a specific variable, keep fixed:
training budget, stopping policy, evaluation procedure, data source.

## Failure Handling

**Simple bugs** (typo, missing import, shape mismatch): fix and re-run.

**Fundamental failures** (OOM, NaN loss, non-convergence): log the failure and
move on to a different idea. Do not keep retrying without a new hypothesis.

## How You Work: Iterative Research Loop

After every experiment, follow this exact cycle:
1. Read the **final score** from the completed run.
2. What happened specifically at **h10** and **h20**?
3. Did position error improve, worsen, or shift across horizons?
4. What concrete failure mode should the **next experiment** target?
5. Create **one** new config based on this analysis.
6. Run that experiment immediately.

Each experiment must be informed by the one before it. This is iterative research.
Decide the next experiment ONLY after seeing the previous result.
After completing a run, immediately start the next one.

## One-Sentence Rule

When in doubt, choose the action that is **most likely to improve the final score
on long-horizon rollouts**, not the action that merely lowers `val_loss`,
shortens training, or increases experiment throughput.
\end{lstlisting}

\subsection{Session Persistence: Native vs.\ Externally-Managed}
\label{sec:appendix-persistence}

The two agents differ in how a session persists across iterations. Claude Code runs continuously for the full $6$-hour wall clock: a single invocation holds its own working state, reads \texttt{summary.tsv} between experiments, and decides what to try next without external orchestration.

Codex's exec mode, by contrast, terminates after a single agentic episode. To equalize the time budget, we wrap it in a restart loop that re-invokes the agent whenever it exits. On each restart, the wrapper reads the cumulative results log (\texttt{summary.tsv}), extracts the best and most recent scores, and prepends a structured continuation prompt that informs the agent of prior progress and instructs it to propose a new hypothesis rather than repeat earlier experiments. The loop runs until the external $6$-hour wall-clock timeout, giving Codex the same total budget as Claude Code.

Both agents therefore receive equal compute time and the same experiment log; they differ only in whether persistence is native (Claude Code) or externally managed (Codex). The externally managed restart loop is the reason a single Codex \emph{trial} in our artifacts contains many \texttt{rollout-*.jsonl} session files -- one per restart -- while a Claude Code trial contains a single session jsonl. 

\clearpage
\section{Evaluation Protocol Details}
\label{sec:appendix-eval-details}

This section elaborates the evaluation protocol of \Cref{sec:eval}: the exact state-transition contract used during open-loop rollout, and the held-out scenario suite.

\subsection{State-Transition Contract}
\label{sec:appendix-contract}

To ensure reproducible evaluation, we specify the exact input/output contract for open-loop rollout in \Cref{tab:transition-contract}.

\begin{table}[H]
\centering
\caption{State-transition contract for open-loop rollout. ``Predicted'' columns indicate model outputs; ``Fed back'' specifies what is used as input at the next rollout step when ground truth is unavailable.}
\label{tab:transition-contract}
\scriptsize
\begin{tabular}{llll}
\toprule
\textbf{Field} & \textbf{Input at $t$} & \textbf{Predicted for $t{+}1$?} & \textbf{Fed back at $t{+}1$} \\
\midrule
Position $(x,y)$    & GT / predicted   & Yes ($\Delta$pos)   & predicted \\
Alive                & GT / predicted   & Yes (logit)         & predicted \\
Velocity $(v_x,v_y)$& GT / predicted   & No (auxiliary only)  & zero \\
Gameplay fields      & GT / predicted   & Yes (masked)        & predicted \\
Terminal             & GT / predicted   & Yes (global)        & predicted \\
\addlinespace
Registry $\mathbf{R}$& ground-truth    & No (static)         & ground-truth \\
Action               & ground-truth     & No (given sequence) & ground-truth \\
Globals              & ground-truth     & No (context)        & last observed \\
Reward               & ground-truth     & No                  & zero \\
\bottomrule
\end{tabular}
\end{table}

At horizon $h{=}1$ (teacher-forced), all inputs are ground-truth.
At $h{>}1$, predicted fields are fed back; non-predicted context (actions, globals) uses ground-truth from the recorded episode.
Velocity is zeroed during rollout because it is not a prediction target---models must infer motion from position history.
This is a deliberate design choice: it tests whether models learn genuine dynamics rather than relying on velocity as a shortcut.

\subsection{Scenario-Based Test Suite}
\label{sec:appendix-scenarios}

Scenario-based tests are controlled micro-experiments that evaluate whether a world model has learned specific game rules, physics, and causal mechanisms---as opposed to merely fitting statistical patterns in trajectory data.
Each scenario constructs an initial game state that isolates a particular interaction, pairs it with a short deterministic action sequence, and checks the model's predictions against the ground-truth outcome for that game rule.
Every scenario generates multiple test cases (typically 20--40) with randomized initial conditions drawn from a fixed seed, ensuring both reproducibility and coverage of geometric variety (e.g., different ball angles, entity positions).
All scenarios share a common validation layer that checks result structure completeness, entity bounds, rollout length matching the action count, and RGB frame integrity.

\Cref{tab:scenario-summary} summarizes the full test suite.  Per-game details follow.

\begin{table}[H]
\centering
\caption{Scenario-based test suite overview. Each scenario group is a distinct game-rule probe; groups provide $20$--$40$ replicate episodes drawn from randomized initial conditions (e.g., varying ball position/velocity or entity placement), so every probe is evaluated over multiple independent instances rather than a single fixed rollout. Counts are auto-generated from \texttt{eval\_results\_scenario\_newformula/results\_base\_model/*.json}.}
\label{tab:scenario-summary}
\small
\begin{tabularx}{\linewidth}{@{}l r r X@{}}
\toprule
\textbf{Game} & \textbf{Groups} & \textbf{Episodes} & \textbf{Key aspects tested} \\
\midrule
\pong        & 15 & 460 & Paddle deflection (static/moving, center/edge, L/R), scoring on miss, wall bounce, game win, same-state/different-actions control \\
\asteroids   & 14 & 420 & Bullet--asteroid hit/split, ship collisions, invincibility, cooldown, bullet wrap/lifetime/pool, wave clear, same-state control \\
\breakout    &  6 & 160 & Ball--paddle deflection, brick hit, paddle miss, wall bounce, level clear, same-state control \\
\snake       &  5 & 160 & Food consumption, wall collision, self-collision, safe move, same-state control \\
\platformer  &  5 & 160 & Coin collection, spike collision, goal reach, safe jump, same-state control \\
\frogger     &  5 & 180 & Car collision, drowning, goal reach, level clear, same-state control \\
\kong        &  5 & 180 & Ladder climb, ladder exit, barrel hit, rescue princess, same-state control \\
\racer       &  5 & 140 & Game over, player avoids enemy, player hits enemy, safe drive, same-state control \\
\midrule
\textbf{Total}  & \textbf{60} & \textbf{1,860} & \\
\bottomrule
\end{tabularx}

\end{table}

\paragraph{\pong (15 scenario groups, 460 episodes).}
Ball--paddle interactions are tested for both paddles under four conditions: center hit with a stationary paddle, center hit with a moving paddle, edge hit with a stationary paddle, and edge hit with a moving paddle (8~groups).
Miss scenarios verify that the opposing player scores when the ball passes either paddle (2~groups).
Edge-miss scenarios confirm that near-miss trajectories just outside the paddle's hit zone still result in a goal (2~groups).
A wall-bounce scenario tests vertical reflection off the top and bottom boundaries.
A game-win scenario verifies terminal conditions when one player reaches the maximum score. A same-state / different-actions control with 180 episodes tests whether the model's predictions vary appropriately as the action changes with the state held fixed.

\paragraph{\asteroids (14 scenario groups, 420 episodes).}
Bullet--asteroid hit scenarios test collision detection and asteroid splitting upon destruction (spawning smaller fragments that increase the entity count).
Ship--asteroid collision scenarios test life-loss and terminal conditions when the ship strikes an asteroid, separately from an \emph{invincible-ship-hit} group that verifies the post-respawn grace window suppresses damage.
Additional scenarios verify the invincibility window wearing off, shoot-cooldown enforcement (\emph{ship-shoot} and \emph{shoot-during-cooldown}), bullet lifetime and despawn, bullet pool exhaustion, toroidal edge wrapping for bullets, and a \emph{safe-thrust} group that confirms thrust alone does not trigger collisions.
A wave-clear scenario confirms termination when all asteroids are destroyed. A same-state / different-actions control with 160 episodes tests whether the model's predictions vary appropriately as the action changes with the state held fixed.

\paragraph{\breakout (6 scenario groups, 160 episodes).}
Ball--paddle deflection tests verify direction reversal and confirm no life is lost on contact.
Brick-hit scenarios check that the struck brick is destroyed and a positive reward is issued.
A paddle-miss scenario confirms life loss when the ball passes below the paddle.
Wall-bounce tests verify reflection off side boundaries.
A level-clear scenario confirms game termination when the last brick is destroyed. A same-state / different-actions control with 60 episodes tests whether the model's predictions vary appropriately as the action changes with the state held fixed.

\paragraph{\snake (5 scenario groups, 160 episodes).}
Food consumption tests verify that the snake's length increases and a new food entity spawns at a different location.
Wall-collision and self-collision scenarios confirm terminal conditions.
Safe-move scenarios ensure the snake traverses a clear path without triggering death. A same-state / different-actions control with 80 episodes tests whether the model's predictions vary appropriately as the action changes with the state held fixed.

\paragraph{\platformer (5 scenario groups, 160 episodes).}
Coin-collection scenarios verify positive reward on pickup.
Spike-collision tests confirm damage or terminal conditions.
Goal-reach scenarios check that reaching the destination triggers the appropriate terminal and reward signals.
Safe-jump scenarios verify that a clear jump arc produces no collision. A same-state / different-actions control with 80 episodes tests whether the model's predictions vary appropriately as the action changes with the state held fixed.

\paragraph{\frogger (5 scenario groups, 180 episodes).}
Car-collision tests verify life loss or terminal conditions when the frog intersects a moving vehicle.
Drowning scenarios test terminal conditions when the frog enters water without log support.
Goal-reach and level-clear scenarios separately test partial success (the frog reaches a destination with lives remaining and resets to the start) and full completion (the final destination triggers game termination). A same-state / different-actions control with 100 episodes tests whether the model's predictions vary appropriately as the action changes with the state held fixed.

\paragraph{\kong (5 scenario groups, 180 episodes).}
Ladder-climb and ladder-exit scenarios verify correct vertical movement between platforms.
A barrel-hit scenario confirms damage or terminal conditions on collision with a rolling barrel.
The rescue-princess scenario tests game termination when the player reaches the final objective. A same-state / different-actions control with 100 episodes tests whether the model's predictions vary appropriately as the action changes with the state held fixed.

\paragraph{\racer (5 scenario groups, 140 episodes).}
A game-over scenario confirms terminal conditions when the player's car is crushed with no remaining lives.
Player-avoids-enemy scenarios verify that dodging an oncoming car increments the score.
Player-hits-enemy scenarios confirm life loss when the car collides with remaining lives $> 0$.
A safe-drive scenario validates non-terminal state during unobstructed driving.

A same-state / different-actions control with 60 episodes tests whether the model's predictions vary appropriately as the action changes with the state held fixed.

\clearpage
\section{Change-Type Classifier Methodology}
\label{sec:supp-classifier}

This appendix specifies the zero-shot classifier referenced in \Cref{sec:results-nontrivial}. It is a per-experiment structured-diff classifier: every experiment is compared against its session's base and assigned one of the nine labels in \Cref{tab:change-type}.

\paragraph{Model and decoding.}
\texttt{gemini-3.1-pro-preview} via the \texttt{google-genai} SDK, temperature $0.0$, one call per experiment ($1{,}335$ calls total). Output is constrained by a JSON response schema with required fields \texttt{label}, \texttt{rationale}, \texttt{secondary\_labels}, \texttt{confidence}, \texttt{is\_novel\_idea}, \texttt{tuning\_changes}, \texttt{structural\_changes}; \texttt{label} and each \texttt{secondary\_labels} entry are restricted to the nine-label enumeration.

\paragraph{Payload.}
Concatenated after the prompt:
\begin{compactenum}
\item unified diff of \texttt{config.json} (base $\to$ experiment);
\item unified diff of \texttt{train.py} (base $\to$ experiment);
\item \texttt{EXPERIMENT.md} (agent's writeup, if provided);
\item composite at $h_1$, $h_{20}$, overall, plus the base's composite;
\item a deterministically pre-parsed list of changed config keys, partitioned into \emph{tuning} (\texttt{lr}, \texttt{batch\_size}, \texttt{dropout}, \texttt{weight\_decay}, \texttt{seed}, \texttt{patience}, \texttt{warmup\_steps}, \texttt{max\_steps}, \texttt{max\_train\_seconds}, \texttt{max\_train\_episodes}, \texttt{eval\_every}, \texttt{label\_smoothing}, \texttt{inference\_iterations}, \texttt{num\_sample\_steps}) and \emph{structural} (everything else, excluding the bookkeeping keys \texttt{run\_id}, \texttt{notes}, \texttt{game\_id}, \texttt{model\_type}, \texttt{cache\_dir}, \texttt{final\_eval\_horizons}).
\end{compactenum}

\medskip
\noindent\textbf{Prompt (verbatim).}
\vspace{2pt}

\begin{lstlisting}[%
  basicstyle=\ttfamily\fontsize{7}{8.4}\selectfont,
  commentstyle={},
  keywordstyle={},
  stringstyle={},
  identifierstyle={},
  emph={}, emphstyle={},
  showstringspaces=false,
  numbers=none,
  breaklines=true,
  breakatwhitespace=true,
  breakindent=0pt,
  postbreak=\mbox{\textcolor{gray}{$\hookrightarrow$}\space},
  frame=single,
  framerule=0.4pt,
  rulecolor=\color{black!40},
  backgroundcolor=\color{backcolor},
  xleftmargin=4pt,
  xrightmargin=4pt,
  linewidth=\linewidth,
  aboveskip=2pt,
  belowskip=2pt,
  keepspaces=true,
  columns=fullflexible,
  upquote=true,
]
You are classifying a machine-learning experiment.

The user will supply:
  - a unified diff of `config.json` (what settings the agent flipped)
  - a unified diff of `train.py` (code changes vs the baseline)
  - optionally `EXPERIMENT.md` (the agent's own write-up)
  - the run's final composite score + the baseline's composite score

Your job is to assign ONE PRIMARY LABEL from this taxonomy:

- HYPERPARAM   : numerical hyperparameters only (lr, batch_size, dropout,
                 patience, warmup_steps, weight_decay, label_smoothing,
                 schedules, inference iterations, seed)
- LOSS         : added or reweighted a training objective term
                 (e.g. position loss weight, alive loss weight,
                 prior/posterior weighting, auxiliary L1 term)
- ARCHITECTURE : changed model capacity or structure -- hidden_dim, depth,
                 window, heads, number of decoder/GRU layers, added or
                 removed an ensemble / latent / decoder path, OR swapped
                 what the model predicts (delta-pos vs absolute-pos vs
                 discrete-pos: representation change IS architecture).
- ROLLOUT      : changed TRAINING-TIME rollout mechanics -- open-loop
                 training, rollout-loss depth, warm-start steps,
                 scheduled-sampling, feedback-state / re-encoded context,
                 detached rollout.
- INFERENCE    : changed ONLY prediction-time behavior without
                 retraining-logic changes -- denoise steps, mask ratio at
                 inference, post-decoding fix, number of masked iterations
                 at eval.
- BUGFIX       : fixed a correctness error -- missing symmetric transform,
                 wrong normalization, coordinate/unit mistake, stray
                 inverse-transform whose pair isn't in the loss.
- DATA_AUG     : added noise / dropout / augmentation on training INPUTS
                 (context_noise_std, state noise, input_noise_std).
- INFRA        : compile flags, AMP, workers, cache, dtype -- performance
                 only, not modeling.
- MULTIPLE     : 3+ distinct levers all actively changed simultaneously
                 with no clear author-stated primary intent.

## Decision procedure -- follow in order

The payload now contains a **pre-parsed list of active config keys**
(deterministic -- computed from the JSON, not your guess). You MUST
USE that list as the source of truth for what changed; do not invent
extra "active" keys from the train.py diff unless they are structural
edits that never reached config.json.

**Step 0. Narrative-name check (run FIRST, before counting).**

Read `EXPERIMENT.md` and the `notes` field.
- If the author names **one specific primary change** (e.g. "label
  smoothing for confidence calibration", "add rollout training on top
  of exp2", "double position loss weight", "no latent + 2-layer GRU"),
  that is the PRIMARY label. Everything else in the diff goes into
  `secondary_labels`.
- If the author names 2 related changes that belong to the same
  category (e.g. "pos_weight=5 + pos_loss_type=l1" -> both LOSS), use
  that category.
- If the author names 0 specific changes, or writes only a non-
  committal note like "best config from exp19" / "seed variance", OR
  gives no EXPERIMENT.md at all, proceed to Step 1.

The narrative check is binding -- prefer the author's stated intent
over your own counting even if the diff looks broad, because
"carryover flags that happen to remain set but weren't the point"
are a real pattern in this dataset.

**Step 1. Partition every *active* change into ONE of two buckets.**

Use the pre-parsed active-keys list in the payload. Add any structural
edits in the train.py diff that never reached config.json but clearly
change the forward pass (e.g. a new nn.Module inserted unconditionally).
Do NOT add entries for parameters that exist only as optional arguments
with defaults matching baseline behavior.

### Bucket A -- `tuning_changes` (scalar knobs, no new behavior)
Only these keys go here. A change here is worth 0 structural points.
- lr, batch_size, dropout, weight_decay, seed
- patience, warmup_steps, max_steps, eval_every
- label_smoothing
- Inference iteration counts (num_sample_steps, inference_iterations)
   when unchanged code simply runs a different number of loops.

### Bucket B -- `structural_changes` (branches code, adds term, etc.)
Every other active change. Includes:
- new weighted loss term or reweighting (LOSS impact)
- new nn.Module, decoder layer, ensemble, head, representation swap
  (ARCHITECTURE impact)
- scheduled sampling, open-loop rollout, detached rollout, feedback-state
  (ROLLOUT impact)
- input noise / state noise / context noise (DATA_AUG impact)
- removing a stray transform whose pair isn't in the loss (BUGFIX)
- physics priors or other hand-crafted forward-pass pathways (ARCHITECTURE)
- changing inference-time numerics in new ways (INFERENCE) -- but ONLY
  if the edit is in `predict_step` / `rollout` / `eval` code paths
  WITHOUT introducing a new code path in the training forward pass.

**Step 2. Apply the gate based on `structural_changes`:**

Let S = len(structural_changes) and K = number of distinct categories
those S changes belong to (LOSS, ARCHITECTURE, ROLLOUT, INFERENCE,
BUGFIX, DATA_AUG).

- If S == 0: label = HYPERPARAM. (A seed + wd + warmup sweep is still
  HYPERPARAM even if it touches three keys -- they're all tuning.)

- If S == 1: label = the one category that change belongs to.

- If S == 2:
    - If both are the same category -> that category.
    - If different categories -> pick the PRIMARY one using the
      author's `notes` / `EXPERIMENT.md`; put the other in
      `secondary_labels`. If author narrative is silent, prefer
      LOSS > ROLLOUT > ARCHITECTURE > DATA_AUG > INFERENCE > BUGFIX.

- If S >= 3 AND K >= 2 AND Step 0 did NOT find a stated primary
  intent: label = MULTIPLE. The experiment touches multiple
  substantive axes at once with no author attribution; the run is
  confounded.

- If S >= 3 AND K >= 2 AND Step 0 DID find a stated primary:
  the narrative wins (already returned in Step 0).

- If S >= 3 AND K == 1: label = that single category (three loss
  reweights is still LOSS; three rollout levers is still ROLLOUT).

**Step 3. Category rules (disambiguate S == 1 or S == 2 cases):**

- *Any new code path in the training forward pass* -- a new nn.Module,
  a new decoder branch, a hand-crafted state-transition pathway -- is
  ARCHITECTURE. It is NOT INFERENCE, even if invoked only from
  predict_step/rollout. INFERENCE is reserved for config-level tweaks
  to existing inference code.

- *Representation change* -- swapping what the model predicts
  (delta-position vs absolute-position, discrete vs continuous,
  symlog-space vs real-space) -- is ARCHITECTURE, even when the loss
  function is simultaneously changed to match. Put LOSS in
  `secondary_labels`.

- *Removing a transform whose pair isn't in the loss* is BUGFIX. A
  planned representation redesign is ARCHITECTURE -- use EXPERIMENT.md
  to disambiguate.

**Step 4. Fill the output fields:**

- `label`: primary category from the taxonomy.
- `rationale`: <= 20 words explaining the primary label.
- `secondary_labels`: up to 2 other categories partially applying.
- `confidence`: "high" only if Step 2 was unambiguous.
- `is_novel_idea`: true if this is a non-obvious domain-specific
  idea (e.g. a platformer physics prior), false for standard ML levers.
- `tuning_changes`: list of Bucket-A keys.
- `structural_changes`: list of Bucket-B changes, each as a short
  string like "added rollout_loss_weight term".
\end{lstlisting}

\clearpage
\section{Session Audit: Reward Hacking and Grey Areas}
\label{sec:appendix-audit}

To verify that agents improved the world model rather than the scorer, we audited all $64$ sessions against the safeguards of \Cref{sec:harness}, reading each session's transcript, \texttt{config.json} and \texttt{train.py} diffs, \texttt{EXPERIMENT.md}, and verifier output.

\paragraph{No scorer tampering.} No session edited the scoring path: \texttt{score.py}, \texttt{run.py}, and the evaluator are byte-identical across all $64$ sessions, and we found no import hijacking or monkeypatching of the scorer. A few sessions edited config keys that would drop evaluation horizons, but those edits are inert---the scorer hard-defaults to horizons $\{1,10,20\}$. Self-reported in-session scores and the independent post-session scores agree closely: the median divergence is $0.008$ over the $61$ sessions with in-session verifier output ($0.003$ against the held-out test split across all $64$). The largest divergences are cases where the independent scorer \emph{exposed} over-optimistic self-evaluation---the safeguard working as intended---rather than cases of the agent inflating its reported score.

\paragraph{Boundary cases.} The audit surfaced a small number of boundary cases, all of which remain positive on the held-out test split.
\begin{compactitem}
    \item \emph{Best-of-seed winners.} A handful of winners are the best seed of an otherwise fixed, structurally modified configuration, selected on validation as the protocol allows. They are real under the independent scorer, with partly seed-sensitive magnitudes (\Cref{sec:supp-seed-robustness}).
    \item \emph{Validation-tuned inference knobs.} A few winners draw part of their gain from inference-time knobs tuned on the validation split---the intended development signal. The search baselines (\Cref{sec:supp-search-baseline}) bound the gain achievable this way.
    \item \emph{A near-frozen-motion winner on \racer.} Because true per-step motion on \racer is close to zero, one winning configuration raises its score by suppressing predicted motion at inference. We read this as eval-aligned calibration rather than a modeling improvement; it nonetheless transfers to the held-out splits ($+0.24$ test, $+0.28$ scenario), reflecting the game's genuinely low-motion regime. This motivates per-game motion-scale normalization as a metric refinement in future releases.
\end{compactitem}
These cases are real under the independent scorer but sit at the boundary of what the benchmark intends to measure. All headline numbers in the paper are held-out test scores rather than in-session values.

\clearpage
\section{Additional Results}
\label{sec:supp-additional-results}

This appendix collects supporting results referenced from \Cref{sec:results}. \Cref{sec:supp-search-baseline} reports the compute-matched hyperparameter-search baselines (random search and Optuna/TPE), and \Cref{sec:supp-seed-robustness} the $96$-session seed-robustness study; \Cref{sec:supp-scenarios} reports per-game breakdowns of the held-out scenario suite; \Cref{sec:supp-best-over-time} shows how each session's best validation score evolves over its $k$ experiments; \Cref{sec:supp-token-efficiency} reports the full token-usage comparison between the two agents summarized at the end of \Cref{sec:results-inter-agent}; \Cref{sec:supp-budget} reports the number of experiments each agent completed per session; \Cref{sec:supp-time-to-best} locates where within a session the winning experiment lands; \Cref{sec:supp-time-distribution} reports per-experiment wall-clock distributions, per-step throughput, and total GPU-hour accounting; \Cref{sec:supp-starter-repro} explains why base-run results differ across agent sessions despite identical code, configuration, and seed; \Cref{sec:supp-valloss} shows per-experiment validation-loss trajectories faceted by base architecture; \Cref{sec:supp-classifier-ablation} reports the change-type classifier's intent ablation and cross-model agreement; \Cref{sec:supp-label-selection} decomposes which change categories over-perform at winning and contains the full nine-label breakdown table; \Cref{sec:supp-case-studies} reports the largest held-out lift per game with the winning mechanism for each; and \Cref{sec:supp-figures} collects the remaining supporting figures that did not fit in the main body.

\subsection{Compute-matched hyperparameter-search baselines}
\label{sec:supp-search-baseline}

We detail here the two search baselines of \Cref{sec:results-controls}. Both---\emph{random search} and Bayesian optimization (\emph{Optuna/TPE}~\citep{akiba2019optuna,bergstra2012random})---optimize the base model's exposed hyperparameters (learning rate, batch size, dropout, weight decay, loss weights, rollout-loss depth, and inference iterations) against the same validation composite ($0.1\,c_1 + 0.2\,c_{10} + 0.7\,c_{20}$) through the identical \texttt{run.py} and \texttt{score.py} the agents used. Each search receives the agents' exact per-cell budget---a $6$-hour session with a $10$-minute per-trial cap (a median of ${\approx}34$ trials per cell)---and, like the agents, is scored on the withheld test and scenario splits only after the search.

\Cref{tab:search-baseline} reports the per-cell best scores on the held-out scenario suite. The agents produce the better model in the large majority of cells: taking each cell's stronger agent session, the best agent beats random search in $27$ of $32$ cells ($1$ loss, $4$ ties; sign-test $p\approx2\times10^{-7}$) and TPE in $25$ of $32$ ($2$ losses, $5$ ties; $p\approx6\times10^{-6}$); on the same-distribution test split the records are $28$--$0$--$4$ and $26$--$2$--$4$. Measured from the shared base, the agents close a median $46\%$ of the remaining headroom to a perfect score, versus $21\%$ for random search (${\approx}2.2\times$; agents ahead on $28$ of $32$ cells).

Two notes on the comparison. First, ``best agent'' takes each cell's better of the two agent sessions, pooling two $6$-hour sessions against one search arm; the single-agent records under the same budget are also clear (on the scenario suite, Codex beats random search $24$--$6$--$2$ and TPE $20$--$7$--$5$, and Claude posts $19$--$6$--$7$ against each). Second, our TPE run sampled every trial from the search space, including the first, so it carries no base-anchored baseline; we therefore report TPE on absolute final performance only, while random search---whose first trial reproduces the base (median gap $0.016$)---carries the headroom figures above. The gap is structural: checking each winning configuration against the base's exposed knobs, none of the $64$ is reproducible by knob settings alone (every winner modified \texttt{train.py}), so knob search reaches the ceiling of the knob space while the agents' edits move that ceiling.

\begin{table}[t]
\centering
\small
\caption{Compute-matched search baselines on the held-out scenario suite, per $(\text{game}, \text{base})$ cell. ``Agent best'' is the stronger of the two agent sessions; ``Random best''/``TPE best'' are each search's best trial. ``HR\%'' is headroom closed, $(\text{best}-\text{base})/(1-\text{base})$, reported for the agents and random search only (TPE carries no valid base anchor---see text). ``Best of 4'' is the absolute winner among Codex, Claude, random, and TPE (tie band $0.005$).}
\label{tab:search-baseline}
\setlength{\tabcolsep}{4pt}
\resizebox{\textwidth}{!}{%
  \begin{tabular}{l l l r r r r r l}
\toprule
Game & Base & Best agent & Agent & Random & TPE & Agent & Random & Best \\
     &      &            & best  & best   & best & HR\%  & HR\%    & of 4 \\
\midrule
\asteroids  & ar-trans. & Claude & 0.658 & 0.628 & 0.643 & $+62$\rlap{$^{\dagger}$} & $+59$ & Claude \\
\asteroids  & d3pm      & Codex  & 0.617 & 0.489 & 0.499 & $+52$ & $+34$ & Codex \\
\asteroids  & dreamer   & Codex  & 0.680 & 0.664 & 0.175 & $+65$\rlap{$^{\dagger}$} & $+63$ & Codex \\
\asteroids  & maskgit   & Claude & 0.534 & 0.475 & 0.476 & $+34$ & $+19$ & Claude \\
\breakout   & ar-trans. & Claude & 0.978 & 0.882 & 0.550 & $+94$ & $+27$ & Claude \\
\breakout   & d3pm      & Codex  & 0.596 & 0.649 & 0.577 & $+4$  & $+24$ & Random \\
\breakout   & dreamer   & Claude & 0.971 & 0.955 & 0.972 & $+64$ & $+52$ & tie \\
\breakout   & maskgit   & Codex  & 0.929 & 0.589 & 0.590 & $+84$ & $+12$ & Codex \\
\frogger    & ar-trans. & Claude & 0.692 & 0.566 & 0.605 & $+60$ & $+5$  & Claude \\
\frogger    & d3pm      & Claude & 0.704 & 0.659 & 0.583 & $+19$ & $+22$ & Claude \\
\frogger    & dreamer   & Codex  & 0.751 & 0.715 & 0.712 & $+29$ & $+21$ & Codex \\
\frogger    & maskgit   & Codex  & 0.899 & 0.677 & 0.663 & $+73$ & $+10$ & Codex \\
\kong       & ar-trans. & Claude & 0.666 & 0.617 & 0.719 & $+32$ & $+23$ & TPE \\
\kong       & d3pm      & Codex  & 0.884 & 0.522 & 0.605 & $+74$ & $+5$  & Codex \\
\kong       & dreamer   & Codex  & 0.853 & 0.846 & 0.800 & $+58$ & $+24$ & Codex \\
\kong       & maskgit   & Claude & 0.602 & 0.511 & 0.583 & $+21$ & $+13$ & Claude \\
\platformer & ar-trans. & Claude & 0.914 & 0.898 & 0.874 & $+77$ & $+74$ & Claude \\
\platformer & d3pm      & Codex  & 0.948 & 0.711 & 0.712 & $+84$ & $+6$  & Codex \\
\platformer & dreamer   & Codex  & 0.925 & 0.918 & 0.926 & $+40$ & $+1$  & tie \\
\platformer & maskgit   & Codex  & 0.687 & 0.692 & 0.696 & $+4$  & $+0$  & tie \\
\pong       & ar-trans. & Claude & 0.880 & 0.877 & 0.881 & $+52$ & $+1$  & tie \\
\pong       & d3pm      & Claude & 0.879 & 0.880 & 0.880 & $+31$ & $+42$ & tie \\
\pong       & dreamer   & Codex  & 0.914 & 0.869 & 0.870 & $+34$ & $-2$  & Codex \\
\pong       & maskgit   & Claude & 0.927 & 0.880 & 0.880 & $+78$ & $+47$ & Claude \\
\racer      & ar-trans. & Codex  & 0.700 & 0.667 & 0.629 & $+40$ & $+35$ & tie \\
\racer      & d3pm      & Claude & 0.600 & 0.588 & 0.598 & $+4$  & $-2$  & tie \\
\racer      & dreamer   & Codex  & 0.727 & 0.695 & 0.670 & $+28$ & $+54$ & Codex \\
\racer      & maskgit   & Claude & 0.576 & 0.577 & 0.568 & $+5$  & $+3$  & tie \\
\snake      & ar-trans. & Codex  & 0.611 & 0.453 & 0.096 & $+57$\rlap{$^{\dagger}$} & $+40$ & Codex \\
\snake      & d3pm      & Codex  & 0.748 & 0.728 & 0.455 & $+5$  & $+3$  & Codex \\
\snake      & dreamer   & Codex  & 0.726 & 0.567 & 0.404 & $+70$\rlap{$^{\dagger}$} & $+51$ & Codex \\
\snake      & maskgit   & Codex  & 0.728 & 0.719 & 0.707 & $+1$  & $-4$  & Codex \\
\midrule
\multicolumn{9}{l}{\footnotesize Best-of-4 (absolute): Codex 14, Claude 8, tie 8, Random 1, TPE 1. Best agent vs.\ random \textbf{27--1--4}} \\
\multicolumn{9}{l}{\footnotesize ($p\approx2\times10^{-7}$), vs.\ TPE \textbf{25--2--5} ($p\approx6\times10^{-6}$). Median headroom closed: agent \textbf{46\%} vs.\ random \textbf{21\%}} \\
\multicolumn{9}{l}{\footnotesize (agent ahead on 28/32 cells). $^{\dagger}$ near-zero base ($<0.15$): headroom is scale-inflated, read with care.} \\
\bottomrule
\end{tabular}
}
\end{table}

\subsection{Seed-robustness study}
\label{sec:supp-seed-robustness}

We detail here the seed-repetition study of \Cref{sec:results-controls}. Repeating all $64$ sessions is prohibitively costly, so we repeat the complete experiment on half the grid---\asteroids, \snake, \frogger, and \racer, all four base architectures, both agents ($32$ cells). For each cell we re-run the entire $6$-hour agent session twice more under new random seeds ($1234$ and $5678$) and combine these with the original paper run, giving three independent sessions per cell and $96$ in total. Each repetition is a full run of the discovery process---the agent starts from the base and must find an improvement again, not merely re-score a checkpoint---and all scores are held-out test composites from the independent evaluator.

\Cref{tab:seed-robustness} reports every cell. The headline is that the \emph{direction} of improvement is robust while the \emph{magnitude} is game-dependent. $94$ of $96$ sessions independently reproduce an improvement; the two exceptions are effectively zero ($-0.003$ and $-0.004$, one seed each of \frogger/Claude/MaskGIT and \snake/Codex/D3PM, both cells our thresholded analysis already labels marginal). The median improvement is $+0.14$, versus a median run-to-run spread of the best scores of $0.049$ (${\approx}2.9\times$), and the improvement exceeds the \emph{largest} seed swing on $24$ of $32$ cells. Improvement is large and seed-stable where the base is weak (\asteroids median $\Delta+0.40$, $8/8$ robust; \snake $+0.27$, $7/8$) and shrinks toward the seed floor where the base is already strong (\racer $+0.08$, only $3/8$ robust). We disclose the most seed-sensitive cells: \snake/Codex/Dreamer reaches best scores of $0.64$/$0.92$/$0.65$ across the three seeds---every repetition improves on its base, but the size varies. This is a half-grid study: strong enough to reject ``seed luck'', not enough to place tight error bars on every cell.

\begin{table}[t]
\centering
\scriptsize
\caption{Seed-robustness study: three independent seeds (the paper run plus $1234$ and $5678$) per cell, held-out test composite. ``$\Delta$'' is $\mathrm{mean(best)}-\mathrm{mean(base)}$ across the three seeds; ``Seed spread'' is $\max-\min$ of the three best-run scores. A cell is seed-robust when $\Delta$ exceeds its seed spread; $8$ cells (starred) do not clear this bar, five of them on \racer where the gains are small.}
\label{tab:seed-robustness}
\setlength{\tabcolsep}{4pt}
\resizebox{\textwidth}{!}{%
  \begin{tabular}{l l l r r r c}
\toprule
Game & Agent & Base & Base (3 seeds) & Best (3 seeds) & $\Delta$ & Seed spread \\
     &       &      & paper/1234/5678 & paper/1234/5678 &        & (max$-$min) \\
\midrule
\asteroids & Codex  & ar-trans. & 0.10/0.10/0.10 & 0.68/0.69/0.68 & $+0.588$ & 0.015 \\
\asteroids & Codex  & d3pm      & 0.30/0.31/0.34 & 0.64/0.44/0.50 & $+0.206$ & 0.201 \\
\asteroids & Codex  & dreamer   & 0.10/0.10/0.10 & 0.71/0.68/0.70 & $+0.596$ & 0.025 \\
\asteroids & Codex  & maskgit   & 0.33/0.37/0.41 & 0.46/0.49/0.46 & $+0.100$ & 0.029 \\
\asteroids & Claude & ar-trans. & 0.10/0.10/0.10 & 0.70/0.69/0.68 & $+0.595$ & 0.016 \\
\asteroids & Claude & d3pm      & 0.32/0.36/0.31 & 0.52/0.50/0.52 & $+0.183$ & 0.019 \\
\asteroids & Claude & dreamer   & 0.09/0.10/0.10 & 0.69/0.70/0.68 & $+0.594$ & 0.012 \\
\asteroids & Claude & maskgit   & 0.38/0.44/0.38 & 0.55/0.46/0.48 & $+0.094$ & 0.085 \\
\snake     & Codex  & ar-trans. & 0.10/0.10/0.10 & 0.83/0.78/0.70 & $+0.674$ & 0.132 \\
\snake     & Codex  & d3pm      & 0.50/0.42/0.48 & 0.54/0.55/0.48 & $+0.052$ & 0.067\rlap{$^{*}$} \\
\snake     & Codex  & dreamer   & 0.10/0.10/0.10 & 0.64/0.92/0.65 & $+0.638$ & 0.284 \\
\snake     & Codex  & maskgit   & 0.43/0.47/0.45 & 0.48/0.50/0.47 & $+0.032$ & 0.031 \\
\snake     & Claude & ar-trans. & 0.10/0.10/0.10 & 0.59/0.54/0.60 & $+0.477$ & 0.064 \\
\snake     & Claude & d3pm      & 0.47/0.42/0.44 & 0.53/0.50/0.50 & $+0.062$ & 0.027 \\
\snake     & Claude & dreamer   & 0.10/0.10/0.10 & 0.61/0.68/0.68 & $+0.558$ & 0.078 \\
\snake     & Claude & maskgit   & 0.50/0.48/0.44 & 0.53/0.52/0.52 & $+0.054$ & 0.008 \\
\frogger   & Codex  & ar-trans. & 0.41/0.23/0.32 & 0.75/0.76/0.77 & $+0.438$ & 0.012 \\
\frogger   & Codex  & d3pm      & 0.64/0.67/0.62 & 0.68/0.70/0.67 & $+0.038$ & 0.032 \\
\frogger   & Codex  & dreamer   & 0.71/0.61/0.69 & 0.78/0.88/0.80 & $+0.151$ & 0.096 \\
\frogger   & Codex  & maskgit   & 0.66/0.67/0.68 & 0.93/0.72/0.81 & $+0.153$ & 0.213\rlap{$^{*}$} \\
\frogger   & Claude & ar-trans. & 0.24/0.35/0.67 & 0.72/0.70/0.73 & $+0.296$ & 0.029 \\
\frogger   & Claude & d3pm      & 0.65/0.67/0.62 & 0.72/0.75/0.78 & $+0.103$ & 0.056 \\
\frogger   & Claude & dreamer   & 0.69/0.67/0.69 & 0.74/0.74/0.79 & $+0.074$ & 0.059 \\
\frogger   & Claude & maskgit   & 0.67/0.67/0.68 & 0.73/0.69/0.68 & $+0.031$ & 0.056\rlap{$^{*}$} \\
\racer     & Codex  & ar-trans. & 0.53/0.67/0.41 & 0.73/0.79/0.76 & $+0.225$ & 0.061 \\
\racer     & Codex  & d3pm      & 0.62/0.63/0.63 & 0.64/0.63/0.63 & $+0.007$ & 0.003 \\
\racer     & Codex  & dreamer   & 0.69/0.67/0.63 & 0.75/0.89/0.75 & $+0.133$ & 0.135\rlap{$^{*}$} \\
\racer     & Codex  & maskgit   & 0.61/0.62/0.61 & 0.63/0.64/0.62 & $+0.014$ & 0.016\rlap{$^{*}$} \\
\racer     & Claude & ar-trans. & 0.50/0.67/0.41 & 0.74/0.78/0.74 & $+0.229$ & 0.042 \\
\racer     & Claude & d3pm      & 0.62/0.63/0.63 & 0.64/0.64/0.77 & $+0.058$ & 0.135\rlap{$^{*}$} \\
\racer     & Claude & dreamer   & 0.68/0.67/0.63 & 0.71/0.83/0.73 & $+0.098$ & 0.116\rlap{$^{*}$} \\
\racer     & Claude & maskgit   & 0.62/0.62/0.61 & 0.64/0.62/0.62 & $+0.013$ & 0.023\rlap{$^{*}$} \\
\midrule
\multicolumn{7}{l}{\footnotesize Per game (median $\Delta$ / median spread / robust cells): \asteroids $+0.397$/$0.022$/$8$--$8$; \snake $+0.270$/$0.065$/$7$--$8$;} \\
\multicolumn{7}{l}{\footnotesize \frogger $+0.127$/$0.056$/$6$--$8$; \racer $+0.078$/$0.052$/$3$--$8$. Overall median $\Delta$ $+0.14$ vs.\ median spread $0.049$;} \\
\multicolumn{7}{l}{\footnotesize $\Delta$ exceeds the largest seed swing on $24/32$ cells. $^{*}$ cell where $\Delta$ does not exceed its seed spread.} \\
\bottomrule
\end{tabular}
}
\end{table}

\clearpage
\subsection{Per-game scenario-suite breakdowns}
\label{sec:supp-scenarios}

Per-game breakdowns of the held-out scenario-suite evaluation (\Cref{sec:appendix-scenarios} defines the suite). Each figure shows the base and agent-best scenario scores on every scenario group defined for that game, with the same $(\text{agent}, \text{base})$ groupings used in \Cref{tab:cell-matrix}.

\begin{figure}[H]
\centering
\includegraphics[width=\linewidth]{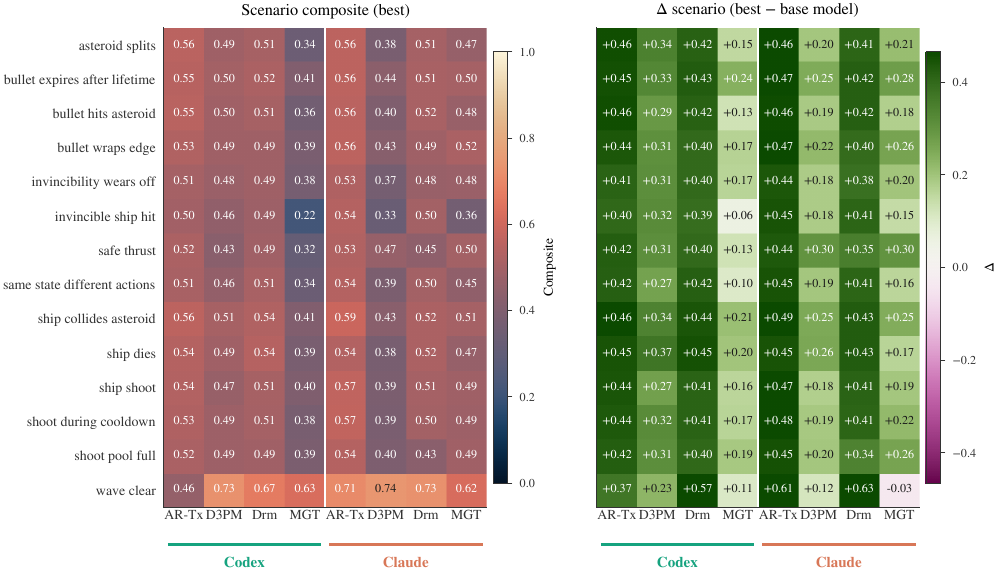}
 \caption{Per-scenario composite ($0.1\,c_1 + 0.2\,c_{10} + 0.7\,c_{h_{\mathrm{end}}}$) and $\Delta$ from base for \asteroids, across the $14$ scenario groups, four base architectures, and two agents. Colour palette matches Table~\ref{tab:cell-matrix}: blue for agent-best composite magnitudes, amber/violet for $\Delta$.}
\label{fig:supp-scenarios-asteroids}
\end{figure}

\begin{figure}[H]
\centering
\includegraphics[width=\linewidth]{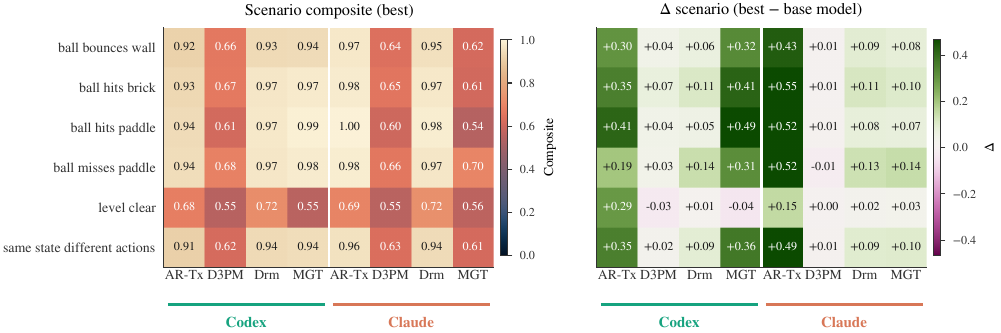}
\caption{Per-scenario composite and $\Delta$ from base for \breakout. Same format as \Cref{fig:supp-scenarios-asteroids}.}
\label{fig:supp-scenarios-breakout}
\end{figure}

\begin{figure}[H]
\centering
\includegraphics[width=\linewidth]{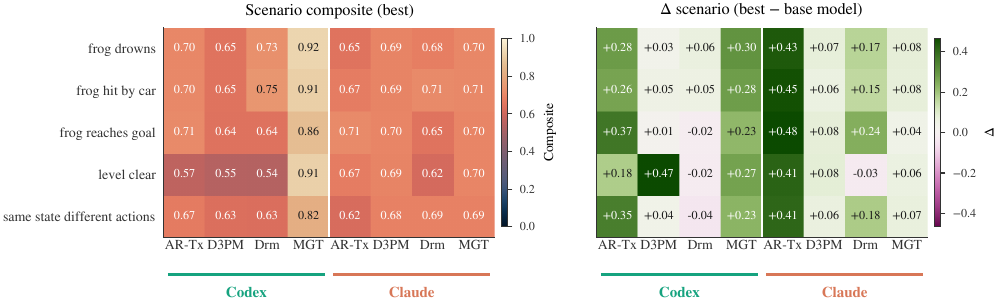}
\caption{Per-scenario composite and $\Delta$ from base for \frogger. Same format as \Cref{fig:supp-scenarios-asteroids}.}
\label{fig:supp-scenarios-frogger}
\end{figure}

\begin{figure}[H]
\centering
\includegraphics[width=\linewidth]{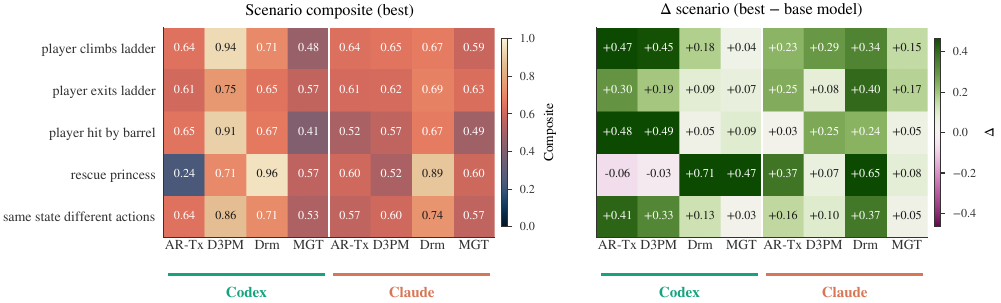}
\caption{Per-scenario composite and $\Delta$ from base for \kong. Same format as \Cref{fig:supp-scenarios-asteroids}.}
\label{fig:supp-scenarios-kong}
\end{figure}

\begin{figure}[H]
\centering
\includegraphics[width=\linewidth]{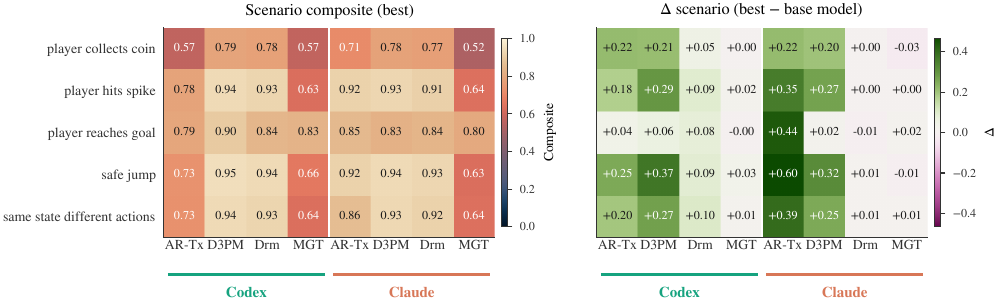}
\caption{Per-scenario composite and $\Delta$ from base for \platformer. Same format as \Cref{fig:supp-scenarios-asteroids}.}
\label{fig:supp-scenarios-platformer}
\end{figure}

\begin{figure}[H]
\centering
\includegraphics[width=\linewidth]{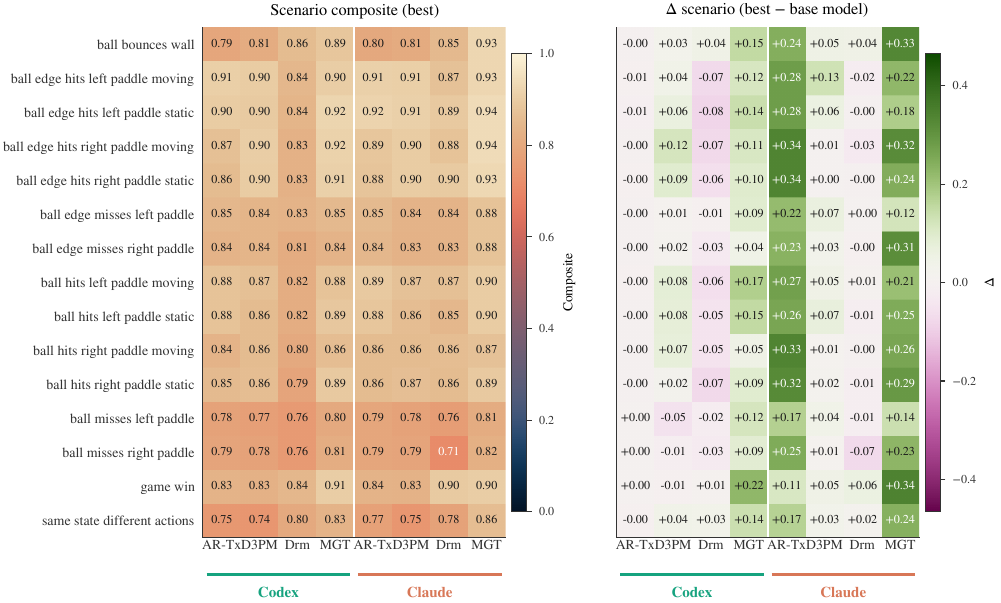}
\caption{Per-scenario composite and $\Delta$ from base for \pong. Same format as \Cref{fig:supp-scenarios-asteroids}.}
\label{fig:supp-scenarios-pong}
\end{figure}

\begin{figure}[H]
\centering
\includegraphics[width=\linewidth]{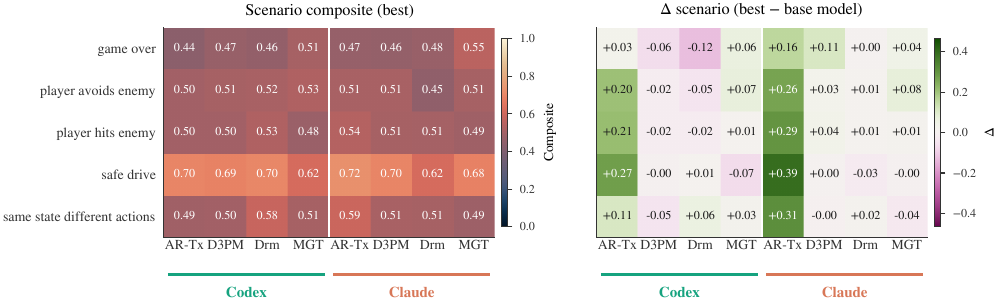}
\caption{Per-scenario composite and $\Delta$ from base for \racer. Same format as \Cref{fig:supp-scenarios-asteroids}.}
\label{fig:supp-scenarios-racer}
\end{figure}

\begin{figure}[H]
\centering
\includegraphics[width=\linewidth]{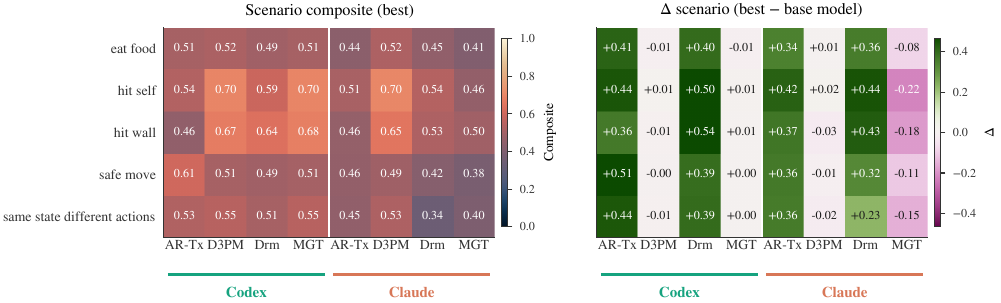}
\caption{Per-scenario composite and $\Delta$ from base for \snake. Same format as \Cref{fig:supp-scenarios-asteroids}.}
\label{fig:supp-scenarios-snake}
\end{figure}

\begin{figure}[H]
\centering
\includegraphics[width=\linewidth]{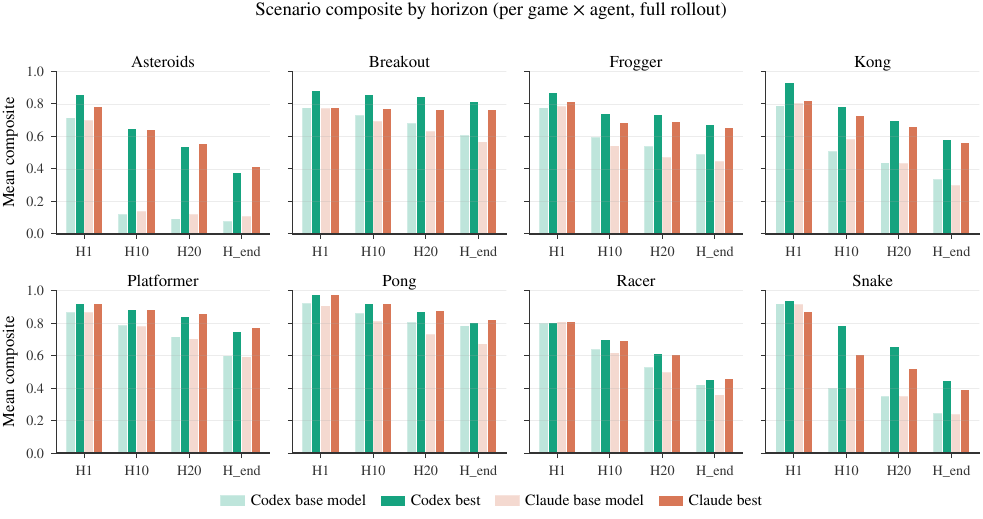}
\caption{Per-horizon scenario composite, split by game (panels) and by agent (colour: Codex green, Opus orange; pale bar = Base, saturated = Agent best), pooled across the four base architectures. Long-horizon lift is most visible on \kong, \platformer, \breakout, and \frogger; \asteroids shows the widest gap at $h_{10}$ before both curves collapse.}
\label{fig:supp-scenarios-horizons}
\end{figure}

\begin{figure}[H]
\centering
\includegraphics[width=\linewidth]{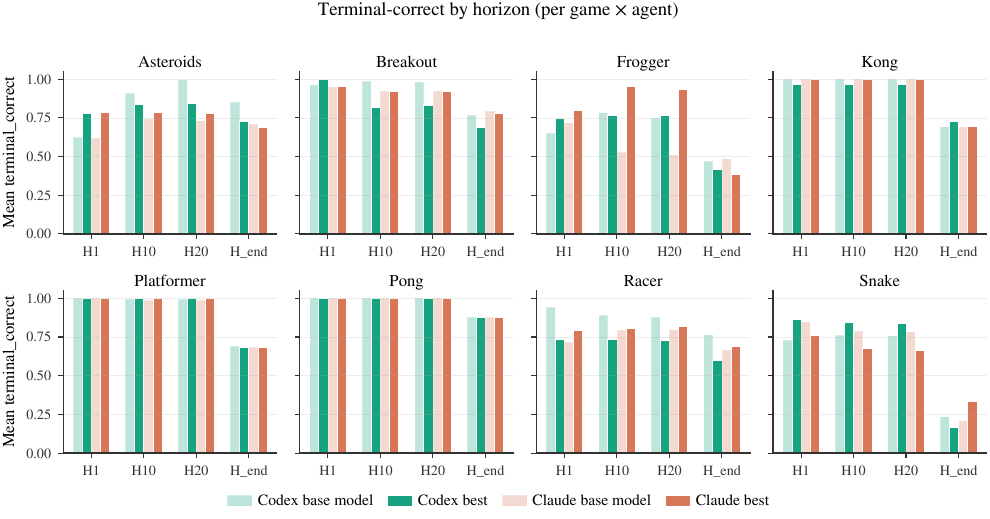}
\caption{\texttt{terminal\_correct} by horizon, split by game and agent. The agent-best advantage is visible at $h_1$ on most games; past $h_{10}$ the gap is either small or reverses. We exclude \texttt{terminal\_correct} from the scenario composite.}
\label{fig:supp-scenarios-terminal}
\end{figure}

\paragraph{Per-horizon decomposition of the scenario score.}\label{par:a2-decomposition}
The scenario score blends three horizon composites: $0.1 \cdot c_1 + 0.2 \cdot c_{10} + 0.7 \cdot c_{h_{\mathrm{end}}}$, where $c_h = 0.9 \cdot (1 - \text{PositionL1}_h) + 0.1 \cdot \text{AliveF1}_h$. \Cref{fig:supp-scenarios-horizons} decomposes the composite at each horizon per (game, agent), pooling across the four base architectures. The long-horizon lift shows up unevenly: on \kong, \platformer and \breakout the agent-best curve stays well above the base through $h_{\mathrm{end}}$; on \pong and \racer both curves remain close; on \asteroids both crash toward zero past $h_{10}$ but the gap at $h_{10}$ is the widest in the suite. An earlier iteration of the composite also included \texttt{terminal\_correct} (whether the model predicts the episode-terminal flag correctly); \Cref{fig:supp-scenarios-terminal} shows that signal is localised to $h_1$ and flattens or inverts past $h_{10}$ on most games, so we exclude it from the scenario composite and include the figure here only as a diagnostic.

\paragraph{Per-horizon decomposition on the scenario split.}
\Cref{tab:scenario-horizon-breakdown} reports the scenario-split counterpart of the test-split \Cref{tab:horizon-breakdown}: per-horizon Base vs.\ Agent-best composite, paired $\Delta$, and the number of sessions (out of 64) with $\Delta > 0$. The pattern matches the test-split story -- small lift at $h_1$, most of the gain at $h_{10}$ and $h_{20}$ -- and extends cleanly to $h_{\mathrm{end}}$ (full curated rollout length): the mean $\Delta$ is $+0.181$ on $55/64$ sessions, confirming that the scenario improvement survives past a truncated $20$-step horizon.

\begin{table}[H]
\centering
\small
\caption{Per-horizon decomposition of the scenario composite (full-rollout mode), aggregated over the $64$ sessions. Cell values are per-horizon composites $c_h$; $\Delta$ is the paired per-session difference. Counterpart to the test-split \Cref{tab:horizon-breakdown}.}
\label{tab:scenario-horizon-breakdown}
\begin{tabular}{l r r r r}
\toprule
Horizon & Base & Agent-best & $\Delta$ & $\#$ sessions with $\Delta > 0$ \\
\midrule
$h = 1$ & 0.821 & 0.874 & $+0.053$ & 50/64 \\
$h = 10$ & 0.576 & 0.768 & $+0.192$ & 60/64 \\
$h = 20$ & 0.507 & 0.710 & $+0.204$ & 60/64 \\
$h_{\mathrm{end}}$ & 0.429 & 0.610 & $+0.181$ & 55/64 \\
\bottomrule
\end{tabular}

\end{table}

\begin{figure}[H]
\centering
\includegraphics[width=\linewidth]{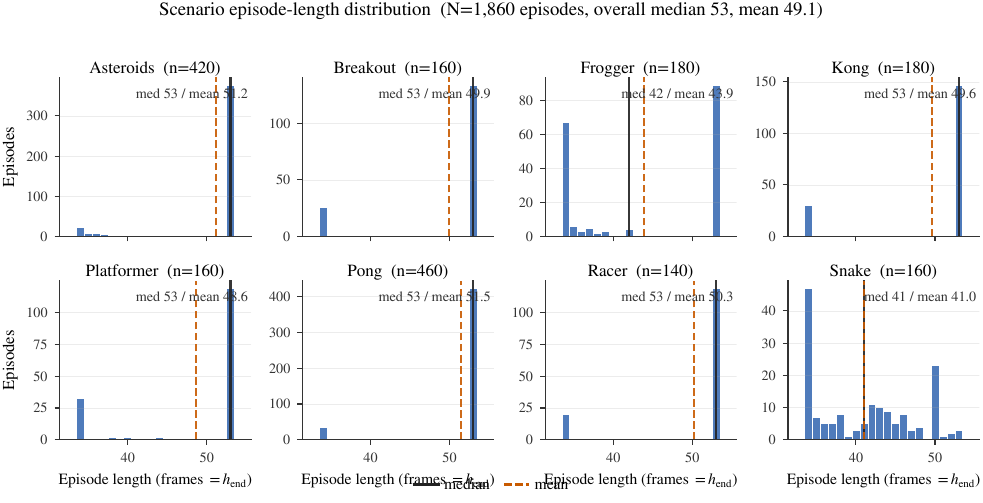}
\caption{Distribution of scenario-episode length (in frames, $=h_{\mathrm{end}}$) per game, across all $1{,}860$ scenario episodes. Black vertical line is the per-game median; orange dashed is the mean. The cap at $53$ frames reflects the scenario construction (pre-event history $32$ ticks + rollout $\le 20$ ticks + one initial tick); \snake and \frogger cases terminate earlier on average because their dynamics end the episode on hit/drown.}
\label{fig:supp-scenarios-episode-length}
\end{figure}

\paragraph{Scenario episode-length distribution.}
The long-horizon term $h_{\mathrm{end}}$ in the scenario score is the full curated rollout length of each episode. That length varies per case because scenarios are built from a fixed pre-event history window plus game-specific post-event rollouts until termination or a cap. \Cref{fig:supp-scenarios-episode-length} shows the empirical distribution of episode lengths across all $1{,}860$ scenario episodes. Most cases reach the $53$-frame cap ($32$ history ticks plus a $20$-tick rollout and a one-tick initial state); games with frequent early termination (\frogger, \snake) have a heavier tail toward shorter rollouts. The per-game spread is small enough that ``$h_{\mathrm{end}}$'' is a well-defined horizon to read per-game scores at, while still being systematically longer than $h_{20}$ on every game.

\subsection{Session progress: best validation score over time}
\label{sec:supp-best-over-time}

\begin{figure}[H]
\centering
\includegraphics[width=\linewidth]{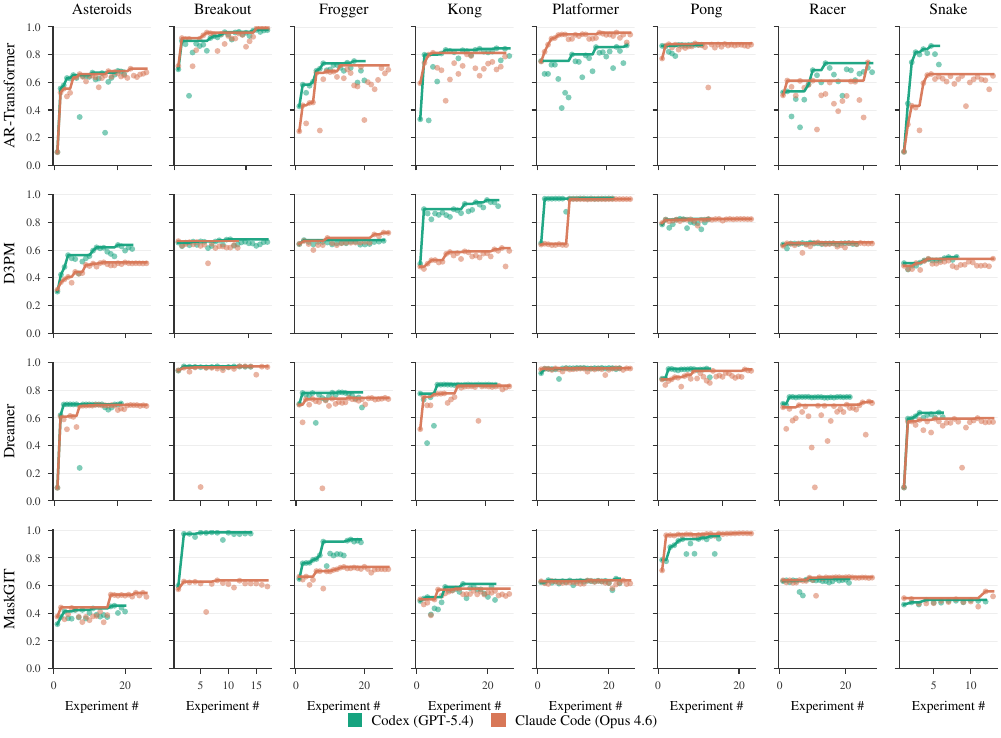}
\caption{Best-so-far validation score over the course of each session, by (game, base). The $x$-axis is the experiment index within the session; lines are step functions that only change when a new best is found. Held-out test scores are computed once at session end on the validation-selected checkpoint.}
\label{fig:supp-best-over-time}
\end{figure}

\Cref{fig:supp-best-over-time} shows how each session's best-so-far validation score evolves across the experiments the agent runs. The $x$-axis is the experiment index within the session; the $y$-axis is the running maximum validation score (the in-session quantity the agent sees) up to that point; held-out test scores are computed once after the session on the validation-selected checkpoint, not per experiment. The layout is an $8 \times 4$ grid of games $\times$ base architectures, with both agents overlaid.

\clearpage
\subsection{Token efficiency}
\label{sec:supp-token-efficiency}

Under the same $6$-hour wall clock and the same task set, \opusfour uses a median $36.9$M prompt tokens per session (mean $38.1$M) versus $25.7$M (mean $23.8$M) for \gptfivefour, a $1.44\times$ ratio. Output tokens: median $354$k (Opus) vs.\ $203$k (Codex), a $1.75\times$ ratio. Cache-read tokens: median $33.5$M (Opus) vs.\ $23.4$M (Codex), a $1.43\times$ ratio. ``Prompt tokens'' counts every input token the model processed in the session, including cache reads, summed across all rollout sessions the harness spawned within the trial. The two providers report cache tokens differently -- OpenAI's \texttt{input\_tokens} already includes \texttt{cached\_input\_tokens}, whereas Anthropic reports \texttt{input\_tokens}, \texttt{cache\_creation\_input\_tokens}, and \texttt{cache\_read\_input\_tokens} as disjoint quantities -- so we normalize both to a single prompt-tokens figure before comparing.

\begin{figure}[H]
\centering
\includegraphics[width=\linewidth]{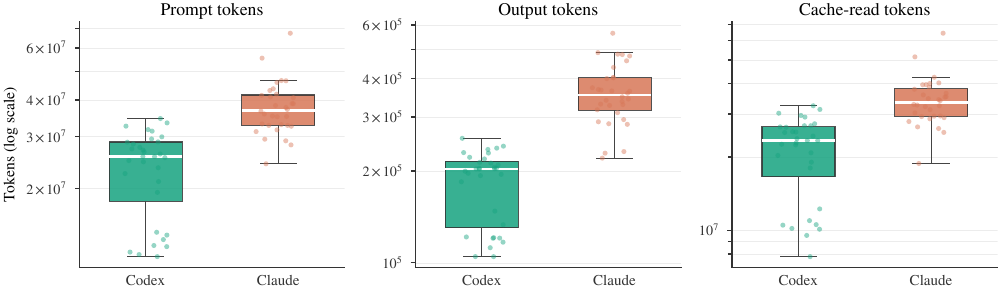}
\caption{Per-session LLM token usage by agent, summed across all rollout sessions within each trial. Boxes indicate the interquartile range, whiskers the $5$th and $95$th percentiles, and jittered points mark individual sessions. All three quantities are shown on log-$y$ axes. Prompt tokens counts every input token the model processed in the session (including cache reads); Codex's \texttt{input\_tokens} already includes cached reads while Opus's does not, so we normalize both to a single prompt-tokens figure. Under the same $6$-hour wall clock, \opusfour consumes roughly $1.4$--$1.8\times$ more tokens than \gptfivefour on each axis.}
\label{fig:token-usage}
\end{figure}

These gaps do not translate into a proportional test-score lift. Summed across the $32$ shared tasks, \opusfour achieves a cumulative $\Delta$ test score of $+5.85$ and \gptfivefour $+6.71$: \gptfivefour produces the larger total improvement despite consuming fewer tokens (median $1.44\times$ fewer total tokens per session). Using the cumulative $\Delta$ and the cumulative token spend per agent, \gptfivefour is approximately $1.65\times$ more token-efficient per unit of score gain on this benchmark.

Two caveats bound this efficiency claim. First, we report harness-reported token counts without pricing; per-token cost for each model at the time of these runs is not available to us, so ``efficient'' here refers to the token budget, not to monetary cost. Second, the two agents differ in both harness behavior (context-window management, tool-use verbosity) and underlying model, so the measurement conflates agent and model. Disentangling these factors would require running \opusfour under the Codex harness and \gptfivefour under Claude Code, which we leave to future work.

\subsection{Experiment budget per session}
\label{sec:supp-budget}

\begin{figure}[H]
\centering
\includegraphics[width=0.85\linewidth]{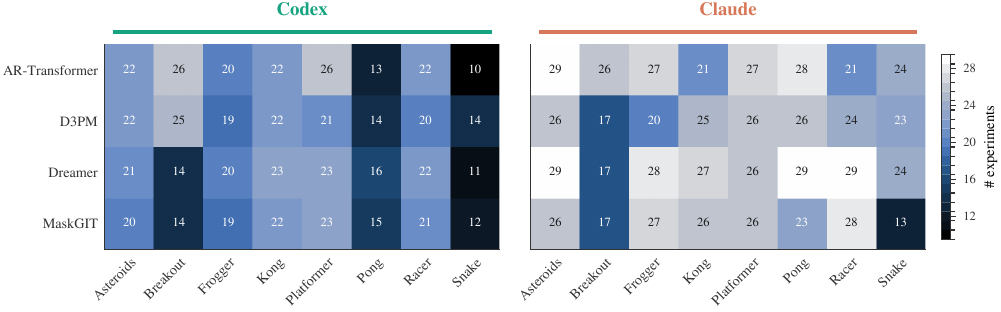}
\caption{Distribution of experiments completed per session, split by agent. Box shows the interquartile range, whiskers the $5$th and $95$th percentiles, and jittered points the $32$ sessions per agent.}
\label{fig:supp-budget}
\end{figure}

\Cref{fig:supp-budget} reports the number of experiments each agent completed per session. \opusfour produces a mean of $23.5$ experiments per session (median $25$) and \gptfivefour $18.2$ (median $19.5$) under the same $6$-hour wall clock; the per-session distribution is summarized by the boxplot and every session is shown as a jittered point. This is the $k$ that enters the best-of-$k$ caveat in \Cref{sec:results-inter-agent}.

\subsection{Time-to-best within a session}
\label{sec:supp-time-to-best}

We index each session's experiments by timestamp and record the position of the winning experiment as a fraction of session length. The median winner occurs at fraction $0.77$ for \opusfour (position $18$ of $25$) and $0.81$ for \gptfivefour (position $11$ of $20$). Approximately half the sessions place their winner in the final $20\%$ of attempts ($47\%$ for \opusfour, $53\%$ for \gptfivefour). This observation is consistent both with continued progress through the session and with a null of ``more draws yield a higher maximum by chance''; disambiguating the two requires a best-at-$k$ analysis at a shared $k$, which is left to future work.

\subsection{Per-experiment wall-clock distribution}
\label{sec:supp-time-distribution}

{Each experiment inside a session runs under a per-experiment wall-clock cap of $600$ s enforced by the harness wrapper (\texttt{elapsed\_seconds} in \texttt{experiments.parquet}). Across the $1{,}335$ non-template experiments, the harness-measured wall clock has median $600.1$ s, $95$th percentile $603.1$ s, and maximum $611.5$ s; the sub-$2$ s overshoots at the tail are attributable to process-shutdown latency rather than real over-runs. $1{,}313$ of $1{,}335$ ($98.4\%$) experiments terminate via the time limit; the rest terminate on the step budget ($21$) or early stopping ($1$). The $10$-minute per-experiment cap therefore held across all experiments.}

{Aggregate wall-clock spent inside training experiments (summing \texttt{elapsed\_seconds}) is $222$ GPU-hours across the $64$ sessions; \Cref{tab:gpu-hour-accounting} reports the per-agent breakdown. The session-level budget is $64 \times 6 = 384$ wall-clock hours, so training experiments occupy roughly $58\%$ of total session time; the remaining $\sim\!42\%$ is agent deliberation, tool use, code editing, and I/O between experiments. \opusshort runs more experiments per session on average ($23.5$ vs $18.2$) but each of its experiments consumes only marginally more training wall-clock, so its total training-time share is higher ($65.2\%$ vs $50.5\%$).}

\begin{table}[H]
\centering
\small
\caption{{Per-agent GPU-hour accounting from \texttt{elapsed\_seconds} in \texttt{experiments.parquet}. ``Training h'' sums per-experiment wall clock over all non-template experiments. ``Budget h'' is the nominal $32 \times 6 = 192$ wall-clock-hour session budget per agent; the remainder goes to agent deliberation, tool use, and I/O.}}
\label{tab:gpu-hour-accounting}
\begin{tabular}{l r r r r}
\toprule
Agent & Experiments & Training h & Budget h & \% in training \\
\midrule
\gptshort & 582 & 97.0 & 192 & 50.5\% \\
\opusshort & 753 & 125.1 & 192 & 65.2\% \\
\midrule
\textbf{Total} & \textbf{1,335} & \textbf{222.1} & \textbf{384} & \textbf{57.8\%} \\
\bottomrule
\end{tabular}

\end{table}

\begin{figure}[H]
\centering
\includegraphics[width=\linewidth]{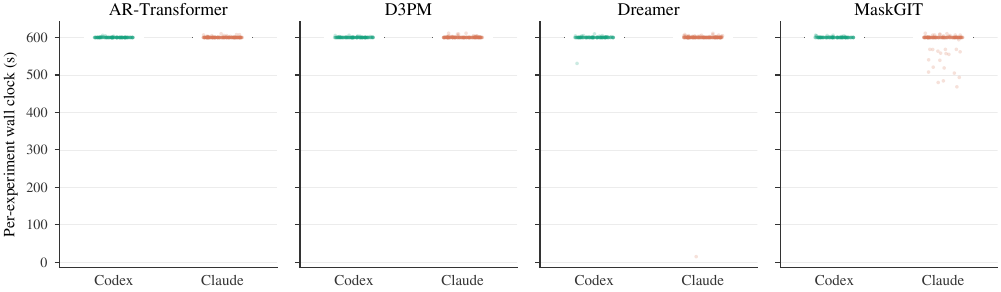}
\caption{{Per-experiment harness-measured wall clock (\texttt{elapsed\_seconds}), split by base architecture and agent. Boxes: IQR; whiskers: $5$th/$95$th percentiles; jittered points: individual experiments. Nearly all experiments run to the $600$ s cap; the thin lower tails correspond to runs that exited early (error or max-steps).}}
\label{fig:suppl-time-by-agent-arch}
\end{figure}

\begin{figure}[H]
\centering
\includegraphics[width=\linewidth]{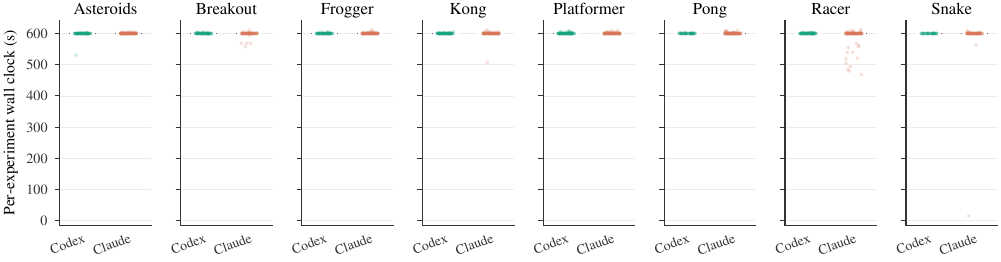}
\caption{{Per-experiment harness-measured wall clock by game, split by agent. Distributions are tight across games because the cap is hard; per-game variation reflects the small fraction of early-terminating experiments.}}
\label{fig:suppl-time-by-game}
\end{figure}

\begin{figure}[H]
\centering
\includegraphics[width=\linewidth]{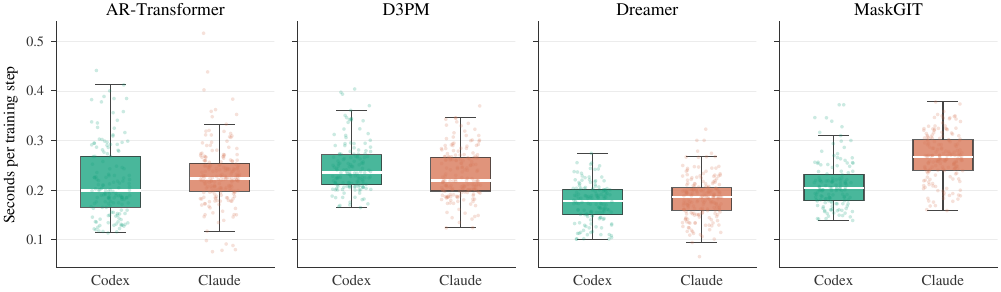}
\caption{{Per-step wall-clock (\texttt{elapsed\_seconds} / \texttt{total\_steps}) by base architecture and agent, over the $1{,}335$ non-template experiments. Dreamer runs fastest per step ($\sim\!0.18$ s), MaskGIT slowest on \opusshort ($\sim\!0.27$ s) due to iterative-decoding eval cost. Systematic per-agent differences on the same base (e.g., \opusshort is faster than \gptshort on Dreamer) reflect per-session compute-environment variation, not code or config.}}
\label{fig:suppl-per-step-time}
\end{figure}

\begin{figure}[H]
\centering
\includegraphics[width=\linewidth]{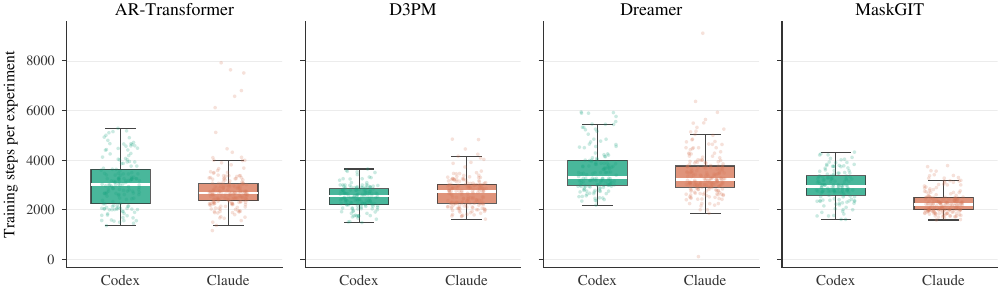}
\caption{{Training steps completed per experiment, by base architecture and agent. Because the wall-clock cap is constant, this distribution is the approximate reciprocal of \Cref{fig:suppl-per-step-time}: faster per-step throughput translates directly into more steps per experiment.}}
\label{fig:suppl-total-steps}
\end{figure}

{\Cref{fig:suppl-time-by-agent-arch} breaks per-experiment wall clock down by base architecture, and \Cref{fig:suppl-time-by-game} breaks it down by game. Because the cap is hard and nearly all experiments run to completion, the distributions cluster tightly near $600$ s with a thin lower tail corresponding to early-terminating runs. \Cref{fig:suppl-per-step-time} reports per-step wall-clock (\texttt{elapsed\_seconds / total\_steps}) by agent and base, and \Cref{fig:suppl-total-steps} shows the corresponding distribution of steps completed per experiment. The in-script \texttt{train\_time\_sec} column reported by \texttt{train.py} over-counts relative to the harness clock (median overcount $87$ s; up to $844$ s on some runs) and is not used for wall-clock accounting here; the figures and aggregates above use the harness-measured \texttt{elapsed\_seconds}.}

\subsection{{Base-run reproducibility across agent sessions}}
\label{sec:supp-starter-repro}

{Each of the $64$ sessions begins by running the unmodified base (\texttt{config\_template}) to establish that session's baseline. Base code, hyperparameters, and fixed seed ($42$) are identical across sessions, so in principle the two agents' base runs on the same (game, architecture) should produce the same model. In practice they do not: step counts differ by up to $2{,}707$ steps on the same task under the same $600$ s wall-clock cap, held-out test composite differs by up to $0.26$, and yet validation loss at the end of training is essentially identical ($|\Delta\,\textrm{val\_loss}| < 0.01$ on $29$ of $32$ pairs and zero gap between \texttt{val\_loss} and \texttt{best\_val\_loss} on every base run).}

\paragraph{{Platform non-determinism.}} {All $64$ sessions ran on a single $8$-GPU node, one session per GPU, with agents and training containers co-scheduled through Docker. Although each session has a dedicated GPU, sessions share CPU cores, host memory bandwidth, PCIe traffic, and disk I/O. Depending on which other sessions were active at the same time, per-step wall clock varies by $\sim\!15\%$ even on the identical base code and configuration (see \Cref{fig:suppl-per-step-time}). Under a fixed $600$ s cap, a faster-running session completes more gradient steps: on \kong/Dreamer, \gptshort's session reached $1{,}634$ steps while \opusshort's session reached $2{,}241$ ($+37\%$). Both runs converged to the same validation loss ($1.3228$ vs $1.3205$), so this is not a matter of one base being ``under-trained''; they simply landed on different points of a plateau that validation loss cannot distinguish.}

\paragraph{{Why Kong/Dreamer diverges despite matched validation loss.}} {The $0.26$ composite gap is driven by long-horizon rollout, not one-step prediction. Per-horizon composites on the held-out test split (\gptshort\,/\,\opusshort): $h_1 = 0.956 / 0.960$ (essentially identical), $h_{10} = 0.842 / 0.661$ (gap $0.181$), $h_{20} = 0.729 / 0.414$ (gap $0.315$). Under one-step prediction the two bases behave the same, and on an i.i.d.\ validation loss that averages short-range terms they look indistinguishable. Divergence appears only under autoregressive open-loop rollout, where small per-step differences in the recurrent state transition compound over the $20$-step horizon. The extra steps \opusshort's base accumulated on Kong/Dreamer moved the weights into a region whose one-step fit is slightly better but whose long-horizon rollout stability is worse---a failure mode of recurrent state-space world models that validation loss does not catch. The other large base-gap tasks in our data (\kong/AR-Transformer, $\Delta = 0.26$; \frogger/AR-Transformer, $\Delta = 0.18$) show the same pattern: matched validation loss, divergence concentrated at $h_{10}$ and $h_{20}$.}

\paragraph{{Consequences and remedies.}} {This is the main reason \Cref{tab:cell-matrix} carries per-agent Base columns rather than a single ``canonical base'' value: the base that \gptshort's session faces differs from the base \opusshort's session faces even when code, config, and seed are identical. Each agent's $\Delta$ test score is therefore computed against its own base, which is the only fair paired comparison. A direct remedy for future releases would be to replace the per-session base with a one-time frozen base checkpoint per (game, architecture). We did not do this here because the per-session base doubles as a runtime sanity check that the agent's environment reproduces the published baseline before any edits are applied; adopting a frozen-base protocol is left to future work.}

\subsection{Per-experiment validation-loss trajectories by base}
\label{sec:supp-valloss}

\begin{figure}[H]
\centering
\includegraphics[width=\linewidth]{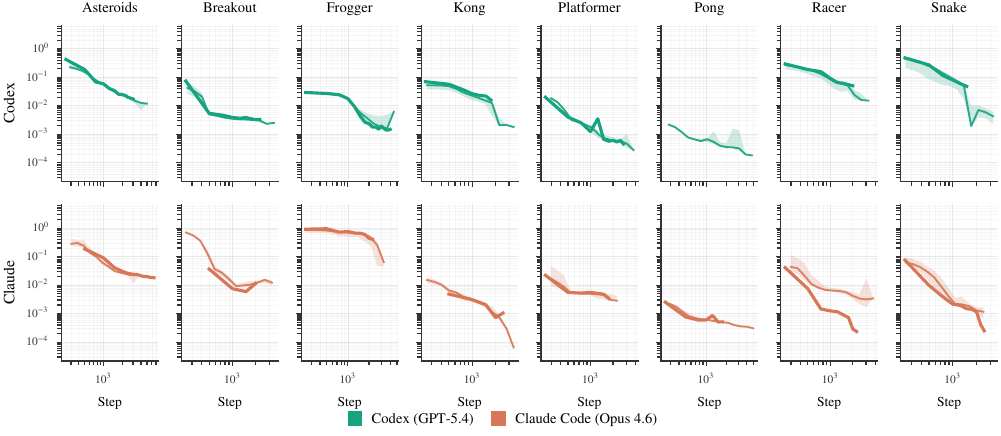}
\caption{Validation-loss trajectories for every AR-Transformer experiment, grouped by game. Each thin line is one agent-produced training run; the thick dashed line is the base.}
\label{fig:suppl-valloss-ar_transformer}
\end{figure}

\begin{figure}[H]
\centering
\includegraphics[width=\linewidth]{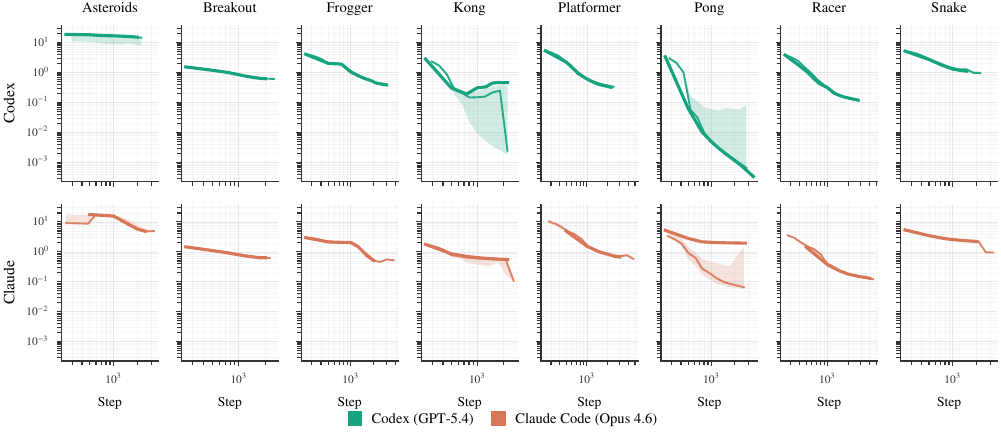}
\caption{Validation-loss trajectories for every D3PM experiment, grouped by game.}
\label{fig:suppl-valloss-d3pm}
\end{figure}

\begin{figure}[H]
\centering
\includegraphics[width=\linewidth]{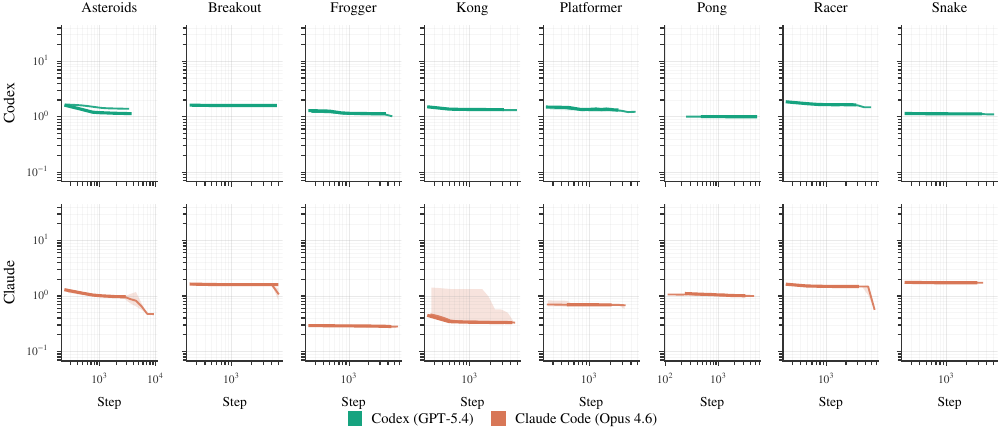}
\caption{Validation-loss trajectories for every Dreamer experiment, grouped by game.}
\label{fig:suppl-valloss-dreamer}
\end{figure}

\begin{figure}[H]
\centering
\includegraphics[width=\linewidth]{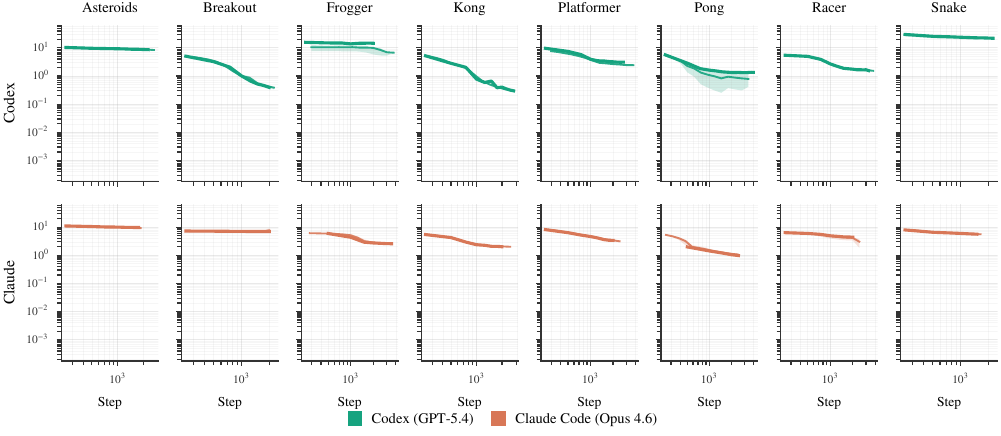}
\caption{Validation-loss trajectories for every MaskGIT experiment, grouped by game.}
\label{fig:suppl-valloss-maskgit}
\end{figure}

\Cref{fig:suppl-valloss-ar_transformer,fig:suppl-valloss-d3pm,fig:suppl-valloss-dreamer,fig:suppl-valloss-maskgit} show validation-loss trajectories over the course of training for every agent experiment, faceted by game within a base architecture. Each panel overlays every $(\text{agent}, \text{experiment})$ training curve on the same axes alongside the base's curve, so the range of optimization trajectories the agents explore for each (game, base) is visible at a glance.

\subsection{Change-type classifier: intent ablation and cross-model agreement}
\label{sec:supp-classifier-ablation}

The change-type labels in \Cref{sec:results-nontrivial} come from a judge that sees the code diff \emph{and} the agent's stated intent (\texttt{EXPERIMENT.md}). To measure how much each label depends on that narrative, we re-ran the classifier on all $1{,}335$ experiments under three input conditions with the prompt held fixed: \textbf{FULL} (the original input), \textbf{FULL2} (an identical rerun, giving a temperature-$0$ noise floor), and \textbf{NOINT} (the agent's stated intent stripped, leaving only the code and config diff). An axis is intent-driven only where FULL-vs-NOINT agreement falls below the FULL-vs-FULL2 floor.

\begin{table}[t]
\centering
\small
\caption{Classifier intent ablation over all $1{,}335$ experiments. Agreement with the original FULL labels under an identical rerun (noise floor) and with the agent's stated intent removed (NOINT).}
\label{tab:classifier-ablation}
\begin{tabular}{l r r}
\toprule
Axis & Noise floor (FULL2) & Intent removed (NOINT) \\
\midrule
Trivial vs.\ non-trivial & $98.8\%$ & $\mathbf{94.6\%}$ \\
Novelty flag             & $96.5\%$ & $89.3\%$ \\
Coarse category          & $96.3\%$ & $56.3\%$ \\
Primary $9$-way label    & $95.0\%$ & $43.1\%$ \\
\bottomrule
\end{tabular}
\end{table}

\Cref{tab:classifier-ablation} shows the split cleanly. The headline trivial/non-trivial axis is near-mechanical: it agrees $94.6\%$ without the narrative, essentially at the noise floor, and the non-trivial \emph{rate} is stable-to-higher when intent is removed ($93.2\%\rightarrow97.7\%$), so \texttt{EXPERIMENT.md} is not inflating it. The fine categories, by contrast, are intent-dependent: without the stated intent the judge cannot identify which change in a multi-change diff was \emph{primary} and retreats to \textsc{multiple} ($16\%\rightarrow63\%$), even though the diff is unchanged (mean structural changes $5.43$ vs.\ $5.37$). We therefore present the fine-grained categories descriptively and anchor the headline on the intent-independent binary. A second- and third-judge check (Claude Opus~4.8 and Claude Haiku~4.5 re-judging a $120$-experiment subsample under both conditions) agrees: every judge$\times$condition places $91$--$99\%$ of the subsample non-trivial. As a preliminary human check, one author re-labeled a random sample of $100$ experiments and matched the judge's trivial/non-trivial call in $90$; a blinded two-annotator pass on this axis is left to future releases.

\subsection{Change-category selectivity}
\label{sec:supp-label-selection}

\Cref{tab:change-type} reports the fraction of each change category that produces a session-best experiment. A complementary measure is the ratio of each category's share of session-winners to its share of the overall experiment population: if an agent allocates $18.8\%$ of its experiments to \textsc{rollout} and $18.8\%$ of its session-winners are \textsc{rollout}, the category is neutral; a larger win-share than population-share indicates selective efficacy. \textsc{inference} exhibits the largest positive selection ratio, accounting for $7.3\%$ of experiments and $15.6\%$ of session-winners ($2.1\times$). \textsc{loss} is similarly over-represented ($18.8\% \rightarrow 28.1\%$, $1.5\times$), and \textsc{hyperparam} is approximately neutral ($8.3\% \rightarrow 9.4\%$, $1.1\times$). \textsc{multiple} is under-represented ($13.8\% \rightarrow 6.2\%$, $0.45\times$), consistent with the interpretation that simultaneously varying multiple axes reduces attributable signal per experiment. \textsc{bugfix} and \textsc{infra} produce no session-bests at the current sample size. Because these are the fine-grained categories, which depend on the agent's stated intent (\Cref{sec:supp-classifier-ablation}), we present this selectivity as descriptive rather than as objective evidence of directed research; the intent-independent trivial/non-trivial split (\Cref{sec:results-nontrivial}) is the claim we rely on.

\begin{table}[H]
\centering
\small
\caption{Breakdown of the $1{,}335$ agent experiments by change type using the nine-label \geminipronew{} taxonomy. ``\% of session-bests'' is the fraction of the $64$ sessions whose session-best experiment carries this label. \textsc{loss}, \textsc{architecture}, \textsc{rollout}, and \textsc{inference} together account for $75\%$ of session-best experiments; \textsc{bugfix} and \textsc{infra} produce no session-bests.}
\label{tab:change-type}
\setlength{\tabcolsep}{5pt}
\begin{tabular}{l r r r r r}
\toprule
Change type & $n$ exps & Mean $\Delta$ & Median $\Delta$ & Win rate & \% of session-bests \\
\midrule
\textsc{architecture} & 248 & $+0.18$ & $+0.14$ & 0.86 & 17\% \\
\textsc{loss} & 251 & $+0.09$ & $+0.04$ & 0.81 & 28\% \\
\textsc{rollout} & 251 & $+0.20$ & $+0.09$ & 0.89 & 14\% \\
\textsc{data\_aug} & 165 & $+0.10$ & $+0.04$ & 0.85 & 9\% \\
\textsc{inference} & 98 & $+0.10$ & $+0.03$ & 0.82 & 16\% \\
\textsc{hyperparam} & 111 & $+0.13$ & $+0.03$ & 0.73 & 9\% \\
\textsc{bugfix} & 23 & $+0.12$ & $+0.03$ & 0.78 & 0\% \\
\textsc{infra} & 4 & $+0.03$ & $+0.03$ & 0.75 & 0\% \\
\textsc{multiple} & 184 & $+0.17$ & $+0.07$ & 0.85 & 6\% \\
\midrule
\textbf{all} & \textbf{1,335} & $\mathbf{+0.14}$ & $\mathbf{+0.06}$ & \textbf{0.84} & \textbf{100\%} \\
\bottomrule
\end{tabular}

\end{table}

\subsection{Case studies: the largest held-out lift per game}
\label{sec:supp-case-studies}

\Cref{tab:case-studies} reports, for each of the eight games, the (agent, base) session with the largest $\Delta$ held-out test score, together with the \geminipronew-assigned label and a short description of the winning mechanism derived from the structured config diff and the agent's \texttt{EXPERIMENT.md} writeup. Seven of the eight winners introduce new training objectives, representation changes, rollout-time procedures, or architectural edits; only \pong/MaskGIT is won by a learning-rate and step-budget schedule.

\begin{table}[H]
\centering
\small
\caption{Largest held-out lift per game. For each of the eight games we report the (agent, base) session with the largest $\Delta$ test score, together with the \geminipronew-assigned label and a short description of the winning mechanism. Seven of the eight winners introduce new training objectives, representation changes, rollout-time procedures, or architectural edits; only \pong/MaskGIT is won by a learning-rate and step-budget schedule.}
\label{tab:case-studies}
\setlength{\tabcolsep}{4pt}
\begin{tabularx}{\linewidth}{l l c c X}
\toprule
Agent & Game / Base & Base$\to$Best & $\Delta$ & Mechanism \\
\midrule
\gptshort & \snake/ar-trans. & 0.10$\to$0.83 & \textbf{$+0.74$} & \textsc{rollout} --- Increasing the rollout tail weight power to penalize late autoregressive drift. \\
\gptshort & \asteroids/dreamer & 0.09$\to$0.71 & \textbf{$+0.61$} & \textsc{architecture} --- A skip connection to feed core state directly to the decoder. \\
\opusshort & \frogger/ar-trans. & 0.24$\to$0.72 & \textbf{$+0.48$} & \textsc{architecture} --- Computing velocity from position differences. \\
\gptshort & \kong/ar-trans. & 0.37$\to$0.84 & \textbf{$+0.48$} & \textsc{rollout} --- Adding a final-step position L1 penalty to the open-loop rollout training. \\
\gptshort & \breakout/maskgit & 0.62$\to$0.98 & \textbf{$+0.36$} & \textsc{data\_aug} --- Modifying the temporal scaling of correlated position drift on context frames. \\
\opusshort & \platformer/d3pm & 0.62$\to$0.97 & \textbf{$+0.35$} & \textsc{inference} --- Removing velocity reconstruction during rollout prediction. \\
\opusshort & \pong/maskgit & 0.68$\to$0.98 & \textbf{$+0.29$} & \textsc{hyperparam} --- Aggressive cosine decay (lr and max\_steps). \\
\opusshort & \racer/ar-trans. & 0.50$\to$0.74 & \textbf{$+0.24$} & \textsc{multiple} --- Alive-gated position loss, zeroed inference scale, and input noise to create a safe baseline. \\
\bottomrule
\end{tabularx}

\end{table}

\paragraph{Recurring mechanism families.} Beyond these eight case studies, reading the winning sessions qualitatively shows the edits converging on a few recurring families---upweighting late rollout steps in the training loss; training under the model's own predictions (scheduled-sampling style); motion priors and state persistence carried across steps; and predicting deltas rather than absolute state. All of these attack a single failure mode, compounding autoregressive error, which is exactly the component the $h_{10}/h_{20}$-weighted score rewards and the one-step objective does not optimize (exposure bias / objective mismatch~\citep{bengio2015scheduled,talvitie2017self,lambert2020objective}). The final winning step is often a tuning refinement of a mechanism the session built earlier.

\subsection{Supplementary figures}
\label{sec:supp-figures}

This subsection reports additional figures covering the full experimental matrix, per-label uplift structure, and the distribution of \geminipronew-flagged novel ideas.

\begin{figure}[H]
\centering
\includegraphics[width=\linewidth]{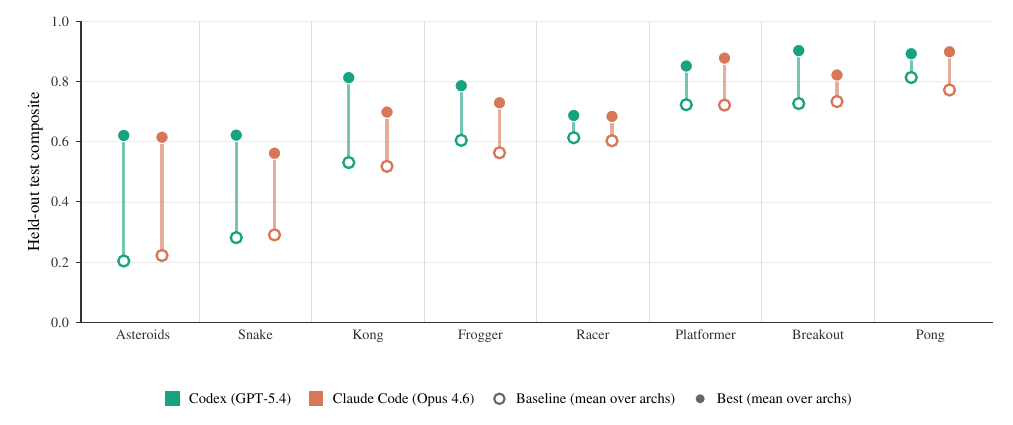}
\caption{Per-game base test score versus agent-best test score. Each vertical dumbbell pairs a game and an agent: the open circle marks the mean base test score across the four base architectures, the filled circle marks the mean agent-best test score, and the segment length encodes the session-best lift. Games are ordered by base strength (hardest on the left). \asteroids and \snake receive the largest lifts; \pong starts near saturation and gains the least.}
\label{fig:game-hardness}
\end{figure}

\begin{figure}[H]
\centering
\includegraphics[width=\linewidth]{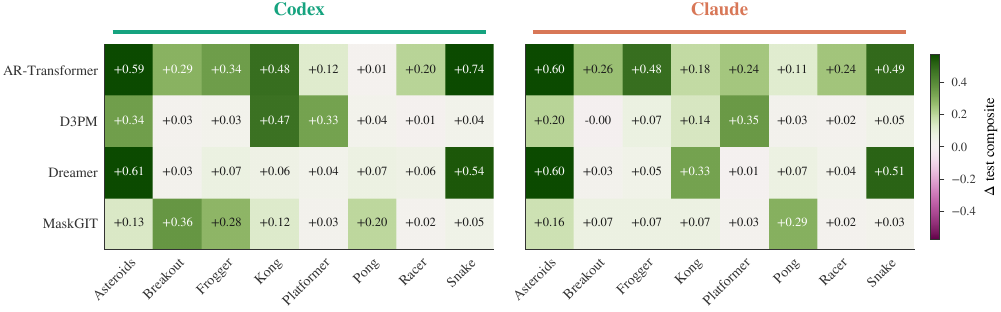}
\caption{Held-out $\Delta$ test score (agent-best minus base) for each of the $32$ tasks, split by agent. Red squares indicate positive lifts; the maximum observed lift is $+0.74$ for \gptfivefour on \snake/AR-Transformer.}
\label{fig:head-to-head}
\end{figure}

\begin{figure}[H]
\centering
\includegraphics[width=\linewidth]{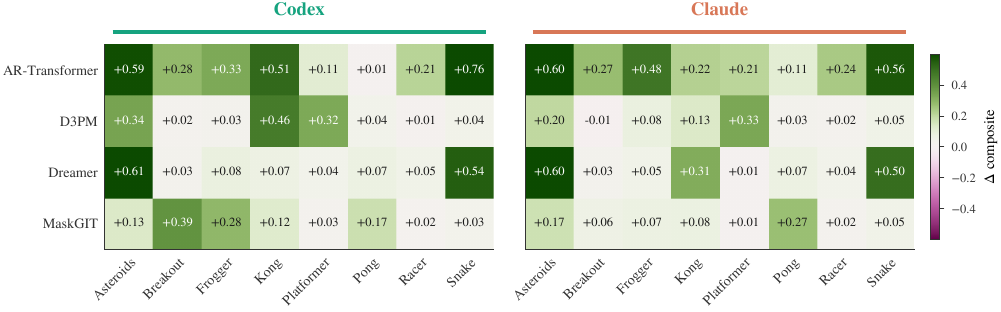}
\caption{Per-task in-session $\Delta$ score (agent-best minus base), split by agent. All $64$ sessions are shown; the single negative square corresponds to {a \opusfour regression} of $-0.006$. This figure complements \Cref{fig:head-to-head} by reporting the in-session measure against which the agent optimised.}
\label{fig:delta-best}
\end{figure}

\begin{figure}[H]
\centering
\includegraphics[width=0.85\linewidth]{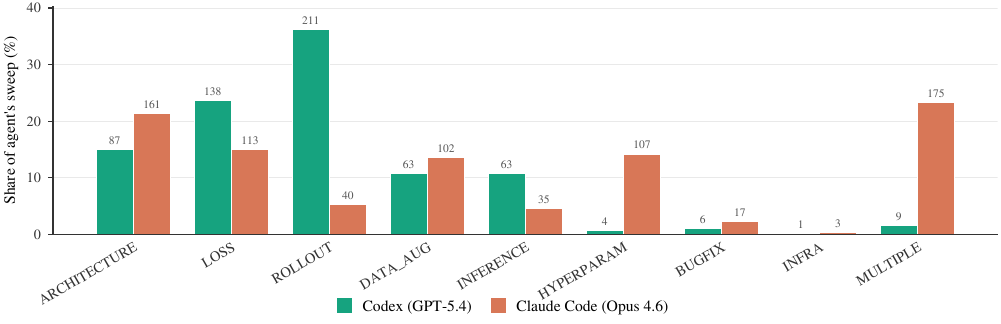}
\caption{Share of each agent's $1{,}335$ experiments across the nine \geminipronew{} change labels. \gptfivefour's experiments concentrate in \textsc{rollout} ($36\%$), \textsc{loss} ($24\%$), and \textsc{architecture} ($15\%$); \opusfour distributes its experiments more broadly across categories, with \textsc{multiple} ($23\%$), \textsc{architecture} ($21\%$), and \textsc{hyperparam} ($14\%$) as its three largest.}
\label{fig:label-mix}
\end{figure}


\begin{figure}[H]
\centering
\includegraphics[width=0.85\linewidth]{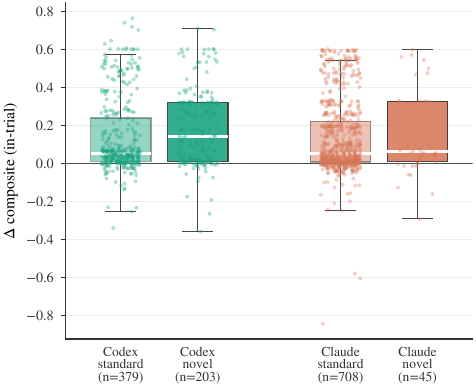}
\caption{Distribution of $\Delta$ in-session score split by the \geminipronew{} \texttt{is\_novel\_idea} flag, faceted by agent. The novelty flag is intent-sensitive (\Cref{sec:supp-classifier-ablation}) and we treat it descriptively rather than as an objective rate; the flagged experiments exhibit a larger upper tail than standard-lever experiments but comparable medians, indicating that novelty does not directly imply a higher expected lift.}
\label{fig:novel-vs-standard}
\end{figure}

\clearpage

\end{document}